\documentclass[11pt]{article}

\usepackage[preprint]{acl}

\usepackage{times}
\usepackage{latexsym}

\usepackage[T1]{fontenc}

\usepackage[utf8]{inputenc}

\usepackage{microtype}

\usepackage{inconsolata}

\usepackage{graphicx}

\usepackage{url}            
\usepackage{booktabs}       
\usepackage{amsfonts}       
\usepackage{tikz}               
\usepackage{multirow}           
\usepackage{enumitem}           
\usepackage{caption}            
\usepackage{pifont}             
\usepackage{subcaption}         
\usepackage{wrapfig}            
\usepackage[edges]{forest}      
\usepackage[most]{tcolorbox}    
\usepackage[table,xcdraw]{xcolor}

\definecolor{softpink}{rgb}{1,0.9,0.9}
\definecolor{softblue}{rgb}{0.9,0.9,1}
\definecolor{softgreen}{rgb}{0.9,1,0.9}
\definecolor{softyellow}{rgb}{1,1,0.9}

\newcommand{\Benchmark}{\texttt{FEANEL}}

\newcommand\Red[1]{\textcolor{red}{#1}}

\title{FEANEL: A Benchmark for Fine-Grained Error Analysis\\in K-12 English Writing}

\author{
    Jingheng Ye$^{1,2}$,
    Shen Wang${^1}$,
    Jiaqi Chen${^2}$,
    Hebin Wang$^{2}$,\\
    \textbf{Deqing Zou}$^{2}$,
    \textbf{Yanyu Zhu}$^{2}$,
    \textbf{Jiwei Tang}$^{2}$,
    \textbf{Hai-Tao Zheng}$^{2}$,\\
    \textbf{Ruitong Liu}${^2}$,
    \textbf{Haoyang Li}$^{1}$,
    \textbf{Yanfeng Wang}$^{3}$,
    \textbf{Qingsong Wen}$^{1}$\\
    $^1$Squirrel Ai Learning,
    $^2$Tsinghua University,
    $^3$Shanghai Jiao Tong University\\
    \texttt{yejh22@mails.tsinghua.edu.cn}
}

\begin{document}
\maketitle

\begin{abstract}
Large Language Models (LLMs) have transformed artificial intelligence, offering profound opportunities for educational applications. However, their ability to provide fine-grained educational feedback for K-12 English writing remains underexplored. In this paper, we challenge the error analysis and pedagogical skills of LLMs by introducing the problem of Fine-grained Error Analysis for English Learners and present the \emph{Fine-grained Error ANalysis for English Learners} (\textbf{\Benchmark{}}) Benchmark.
The benchmark comprises 1,000 essays written by elementary and secondary school students, and a well-developed English writing error taxonomy.
Each error is annotated by language education experts and categorized by type, severity, and explanatory feedback, using a part-of-speech-based taxonomy they co-developed.
We evaluate state-of-the-art LLMs on the \Benchmark{} Benchmark to explore their error analysis and pedagogical abilities. Experimental results reveal significant gaps in current LLMs' ability to perform fine-grained error analysis, highlighting the need for advancements in particular methods for educational applications\footnote{Dataset is available at \url{https://huggingface.co/datasets/Feanel/FEANEL}.}. 

\end{abstract}

\section{Introduction}
\label{sec:introduction}
Large Language Models (LLMs) have revolutionized artificial intelligence with their extensive knowledge and remarkable reasoning capabilities~\citep{he2025survey}, creating unprecedented opportunities in educational applications~\citep{wang2024large,yan2024practical,chu2025llm,xu2024large}. In language education, LLM-powered solutions are increasingly being deployed to enhance personalized learning experiences~\citep{ye2025position,liu2025erevise+}. However, while LLMs demonstrate impressive performance in many tasks, their application in providing fine-grained educational feedback targeted at each error humans may make~\citep{wang-etal-2024-bridging,stahl-etal-2024-exploring,yan2024errorradar,han2023fabric}, which is critical for language acquisition, remains under-explored.

Current related methodologies, however, typically focus on surface-level corrections~\citep{bryant2023grammatical,huang-etal-2023-frustratingly,ye-etal-2023-cleme} or global assessments with coarse feedback~\citep{do-etal-2024-autoregressive-multi,ke2019automated}, which do not capture the multifaceted nature of writing difficulties. Moreover, the lack of a standardized taxonomy for English writing errors~\citep{zou2025revisiting} has led to inconsistencies in error categorization and hindered the development of robust educational tools. These gaps are particularly pronounced in K-12 English education, where learners exhibit diverse proficiency levels and error patterns that require fine-grained analysis and personalized feedback.

Therefore, this paper define the problem of \emph{Fine-grained Error Analysis for Language Learners}, a crucial component of language education aimed at systematically analyzing learners' errors in written English. Error analysis, as a foundational methodology in second language acquisition research~\citep{james2013errors,erdougan2005contribution}, serves two primary purposes: (1) investigating the underlying causes of errors to facilitate targeted interventions, and (2) providing insights into common difficulties in language learning to inform teaching practices and materials. By offering detailed feedback on error types, severity, and corrections, this problem not only supports learners in scaffolding their knowledge but also enhances their ability to learn from mistakes through instant and interpretable feedback~\citep{daheim-etal-2024-stepwise,ye2025excgec}.

To investigate this problem, we introduce the \emph{Fine-grained Error ANalysis for English Learners} (\textbf{\Benchmark{}}) Benchmark, designed to advance research in fine-grained error analysis. The benchmark includes a dataset of 1,000 essays written by K-12 students, with 500 essays (3,003 errors) from elementary school students and 500 essays (5,671 errors) from secondary school students, covering a wide range of age groups and proficiency levels. Each error analysis has been meticulously annotated with an error type, severity level, and explanation, guided by a taxonomy co-developed with language education experts.

\begin{figure*}[tbp!]
    \centering
    \includegraphics[width=2.0\columnwidth]{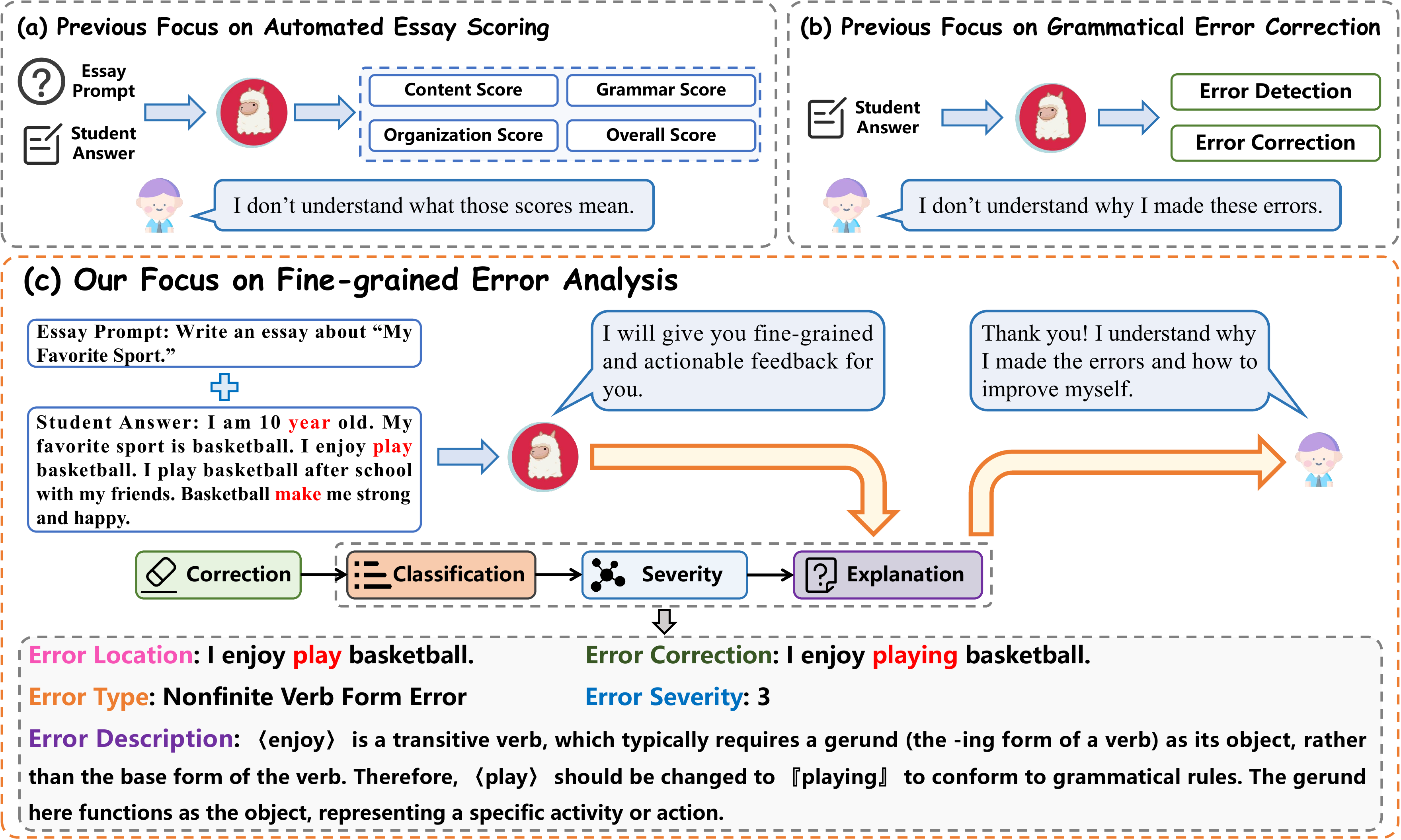}
    \caption{Comparison of our focus on Fine-grained Error Analysis with existing studies.}
    \label{Fig:Intro}
\end{figure*}

As illustrated in Figure~\ref{Fig:Intro}, this paper goes beyond conventional automated essay scoring~\cite{ke2019automated} and grammatical error correction~\citep{bryant2023grammatical,zeng-etal-2024-evaluating} by highlighting the interpretability and educational value of feedback, thereby facilitating a more thorough evaluation of LLMs within language education. Moreover, the benchmark establishes a rigorous framework for evaluating fundamental competencies of LLMs, including their comprehension of syntactic, grammatical, and lexical knowledge and their capacity to replicate pedagogical scenarios by producing engaging and didactically significant feedback. Additionally, \Benchmark{} also evaluates LLMs' capacity for commonsense reasoning and knowledge application, recognizing that effective error analysis in essays often involves understanding logical relationships and world knowledge (e.g., an essay requiring an introduction of tourist attractions in Beijing). This multidimensional evaluation provides a more detailed understanding of LLMs' efficacy in providing insightful and contextually relevant feedback to learners.

We conduct extensive experiments to evaluate the performance of various LLMs on \Benchmark{}. Our empirical study reveals:
(1) LLMs still face significant challenges in classifying complex errors, and often fall short of human-level pedagogical nuance in their explanations.
(2) Performance of LLMs is highly dependent on the detail level of prompts and the availability of examples.
(3) LLMs are sensitive to the sub-task execution order.
We believe that our proposed \Benchmark{} and findings are crucial for educational applications and understanding LLMs' pedagogical ability. In summary, our contributions are threefold:
\begin{itemize}[leftmargin=*]
\item We define the problem of Fine-grained Error Analysis and introduce the \Benchmark{} Benchmark, a novel dataset annotated by English education experts for fine-grained error analysis in writing.

\item We develop a well-defined and part-of-speech-driven taxonomy for English writing errors, addressing the issues of inconsistent categorization and insufficient granularity in previous work.

\item We conduct a comprehensive empirical evaluation of various LLMs, providing insights into their capabilities and limitations in generating interpretable and pedagogically valuable feedback.
\end{itemize}

\section{The \Benchmark{} Benchmark}
\label{sec:benchmark}

\subsection{Problem Definition}
Given an essay prompt $P$, a student's written answer $X$, its corrected version $Y$, and a set of edits $\mathbb E=\{e_1,e_2,\cdots,e_N\}$ that transform $X$ to $Y$\footnote{In this paper, each edit $e_i\in \mathbb E$ is considered an instance of an error, as edits are introduced only to rectify mistakes.}, the problem of fine-grained error analysis focuses on analyzing each specific edit/error. Each analysis comprises three key elements: (1) \textbf{Error Classification}: Categorize the error into an error type $t_i\in\mathcal{T}$ based on the pre-defined taxonomy $\mathcal{T}$. (2) \textbf{Error Severity Rating}: Assign a numerical score $s_i\in\{1,2,3,4,5\}$ to indicate how critically each error affects the sentence’s overall structure and meaning. (3) \textbf{Error Explanation}: Provide an accurate, relevant, and sufficient explanation $d_i$ of why it is an error and how to correct or prevent it. Notice that there may be multiple errors in an edit. We require LLMs to assign a single error type with the highest priority, while also explaining all error types in the explanation. By default, LLMs are required to generate an error analysis in the order of Error Severity $\to$ Error Type $\to$ Error Explanation, which is formulated as:

\vspace{-5pt}
\begin{small}\begin{equation}\begin{aligned}
& P(s_i,t_i,d_i \mid X,Y, e_i) = P(s_i \mid X, Y, e_i) \cdot \\
& ~~~~ P(t_i \mid X, Y, e_i, s_i) \cdot P(d_i \mid X, Y, e_i, s_i, t_i)
\end{aligned}\end{equation}\end{small}

\subsection{Dataset Construction}

\paragraph{Data Collection.}
We collected original essays from two distinct sources to ensure diversity in learner proficiency and educational context, which offers a rich spectrum of fine-grained error analysis for varying writing capabilities. First, essays from elementary school students aged 9-11 were collected through a global online education platform, where students are primarily non-native speakers from around the world. Other essays were sourced from the TECCL Corpus\footnote{\url{https://corpus.bfsu.edu.cn/info/1070/1449.htm}}, a significant corpus of Chinese EFL learners' writing. It is notable for its wide array of over 1,000 essay topics and its representation of learners from elementary to postgraduate levels. For our study, we selected essays corresponding to middle school proficiency (corresponding to students aged 12-18). These essays primarily cover topics such as family, study, friendship, and life. All data from both sources was fully anonymized before use to ensure compliance with privacy regulations. We remove the essays concerning personal privacy.


\paragraph{Data Cleaning.}
To ensure the quality and relevance of the data for our analysis, a rigorous cleaning process was implemented by a team of annotators. Essays were excluded if they: (1) significantly deviated from the given prompt, rendering them off-topic; (2) contained an insufficient number of words, indicating a lack of substantive response; (3) were entirely error-free, as our focus is on error analysis; or (4) were incomplete or nonsensical. Following this filtering, the original formatting of the retained essays was preserved to maintain the authenticity of student responses with real errors. Finally, 500 elementary and 500 secondary essays are selected to construct the dataset.

\paragraph{Data Annotation.}

The data underwent annotation by English education experts with over five years of teaching experience. The comprehensive annotation workflow was as follows:
\begin{itemize}[leftmargin=*]
\item [(1)] \textbf{Error Detection and Correction}: Following established Grammatical Error Correction (GEC) guidelines~\citep{bryant2023grammatical}, education experts rewrote student essays, applying the \textit{minimal} necessary corrections to preserve the original meaning of the essays. This principle ensures objectivity and facilitates accurate error categorization. Experts subsequently reviewed these corrections, checking for over-corrections, missed corrections, or incorrect revisions to guarantee accuracy and consistency. We then leverage the edit extraction tool \textit{CLEME}~\citep{ye-etal-2023-cleme,ye2024cleme2} to extract a set of edits describing the revision trajectory from $X$ to $Y$ for subsequent annotation of error analysis.

\item [(2)] \textbf{Error Type Taxonomy}: In collaboration with two experienced educators, we developed an error type taxonomy comprising 29 distinct error types, primarily based on part-of-speech categories (see Appendix~\ref{app:taxonomy}). The taxonomy was designed separately from the dataset for generalization, highlighting \textit{broad coverage} and \textit{mutual exclusivity} with appropriate granularity. It enables precise classification of nearly all errors without resorting to a vague ``Other Error'' category. Before formal annotation, the taxonomy underwent evaluation and enhancement utilizing a subset of the dataset. A prioritization system is detailed in Appendix~\ref{app:taxonomy}, designed to ascertain the most significant error type in cases of complex errors that involve multiple error types.

\item [(3)] \textbf{Error Analysis}: Each identified error underwent a three-step analysis: (1) \textit{Error Classification}: Experts assigned the most prioritized error type to each edit. (2) \textit{Error Severity Rating}: A 1-5 scale was used to determine the seriousness of each error's impact on sentence structure and meaning. We define and exemplify each severity level in Appendix~\ref{app:severity}. (3) \textit{Error Explanation}: At least two experts independently provided detailed explanations for each error, adhering to principles of accuracy, relevance, and sufficiency. Another expert then reviewed and selected the most appropriate explanation, refining it for clarity and completeness if necessary.
\end{itemize}



\subsection{Dataset Analysis}

\begin{table}[tbp!]
\renewcommand{\arraystretch}{1.0}
\renewcommand{\tabcolsep}{3pt}
\centering
\resizebox{\linewidth}{!}{
\begin{tabular}{lccccc}
    \toprule

    & \textbf{Essays} & \textbf{Essay Len.} & \textbf{Edits/Essay} & \textbf{Edit Len.} & \textbf{Exp. Len.} \\

    \midrule

    \textbf{Ele.} & 500 & 48.2 & 6.01 & 1.66 & 69.27 \\

    \textbf{Sec.}  & 500 & 127.1 & 11.34 & 1.53 & 52.79 \\

    \bottomrule
\end{tabular}}
\caption{Dataset statistics of the \Benchmark{} benchmark. We list the average length and edit number per essay. Length (Len.) is presented as word count.}
\label{Tab:statistics}
\end{table}

\paragraph{Dataset Statistics.}
The overall statistics of the \Benchmark{} benchmark are presented in Table~\ref{Tab:statistics}. The dataset comprises 1,000 essays, evenly split with 500 from elementary school students and 500 from secondary school students. Secondary school essays are considerably longer. This difference in length correlates with the number of identified edits: elementary school essays contain 3,005 edits, while secondary school essays feature significantly more, 5,671 edits. In total, the benchmark includes 8,676 fine-grained error analyses. We provide further analysis of error type distribution on both domains in Appendix~\ref{app:error_type_distribution}.

\subsection{Evaluation Metrics}
We evaluate the fine-grained error analysis task along its three core components. For Error Classification, we report Accuracy to capture overall correctness and Macro-F1 to ensure balanced assessment across all categories, particularly highlighting performance on long-tail distributions. Error Severity Rating is assessed using Mean Absolute Error (MAE). To evaluate Error Explanation quality, we employ standard n-gram-based metrics, i.e., BLEU~\cite{papineni-etal-2002-bleu}, METEOR~\cite{banerjee-lavie-2005-meteor}, and ROUGE-L~\cite{lin2004rouge}. This combined set of metrics provides a robust and complementary evaluation protocol.


\section{Experiments}
\label{sec:experiments}

\subsection{Experimental Settings}
\label{subsec:exp_settings}

\paragraph{Baseline Models.} We evaluate the performance of various LLMs on our \Benchmark{} benchmark to investigate their ability to perform error analysis. The experiment involve state-of-the-art reasoning models as well as other representative models, including GPT-4o~\citep{hurst2024gpt}, o1~\citep{jaech2024openai}, o3~\citep{gpto4}, o4-mini~\citep{gpto4}, Gemini-2.5-pro~\cite{team2023gemini}, DeepSeek-R1~\citep{guo2025deepseek}, Claude-3.7-Sonnet~\citep{Claude37}, Claude-3.7-Sonnet-Thinking~\citep{Claude37}, Grok-3-Beta~\citep{grok}, Qwen-3~\citep{qwen3}, Llama-3~\citep{grattafiori2024llama}, and Mistral-Small-3.1~\citep{mistral-small}. These models represent a mix of closed-source commercial systems and open-source models, ensuring a comprehensive evaluation across different architectures and training paradigms. We report evaluation details and design prompts in Appendix~\ref{app:evaluation_details}.

\begin{table*}[tb!]
\centering
\renewcommand{\arraystretch}{1.1}
\renewcommand{\tabcolsep}{3pt}
\resizebox{0.89\linewidth}{!}{
\begin{tabular}{lccccccccccccc}
\toprule

\multirow{2}{*}{\textbf{Model}} & \multirow{2}{*}{\textbf{Think}}


& \multicolumn{4}{c}{\textbf{Classification}}
& \multicolumn{2}{c}{\textbf{Severity}}
& \multicolumn{6}{c}{\textbf{Explanation}}

\\

\cmidrule(lr){3-6} \cmidrule(lr){7-8} \cmidrule(ll){9-14}


&
& \multicolumn{2}{c}{\textbf{Acc}$\uparrow$}
& \multicolumn{2}{c}{\textbf{F$_1$}$\uparrow$}
& \multicolumn{2}{c}{\textbf{MAE}$\downarrow$}
& \multicolumn{2}{c}{\textbf{BLEU}$\uparrow$} 
& \multicolumn{2}{c}{\textbf{METEOR}$\uparrow$}
& \multicolumn{2}{c}{\textbf{ROUGE}$\uparrow$} \\

\midrule

\textbf{GPT-4o} & \ding{55} & 61.74 & \cellcolor{softgreen} 66.79 & 46.55 & \cellcolor{softgreen} 52.67 & 0.87 & \cellcolor{softgreen} 0.77 & 1.35 & \cellcolor{softgreen} 1.19 & 17.87 & \cellcolor{softgreen} 17.58 & 24.29 & \cellcolor{softgreen} 23.02 \\

\textbf{o1} & \ding{51} & 68.60 & \cellcolor{softgreen} 74.27 & \bf 
62.64 & \cellcolor{softgreen} 64.50 & 0.68 & \cellcolor{softgreen} 0.60 & 0.82 & \cellcolor{softgreen} 1.00 & 15.37 & \cellcolor{softgreen} 16.12 & 23.95 & \cellcolor{softgreen} 23.04 \\

\textbf{o3-low} & \ding{51} & 70.93 & \cellcolor{softgreen} 74.79 & 62.47 & \cellcolor{softgreen} \bf 66.06 & 0.69 & \cellcolor{softgreen} 0.63 & 1.32 & \cellcolor{softgreen} 1.16 & 17.62 & \cellcolor{softgreen} 17.10 & 24.12 & \cellcolor{softgreen} 22.95 \\

\textbf{o4-mini-low} & \ding{51} & 69.80 & \cellcolor{softgreen} 71.43 & 55.28 & \cellcolor{softgreen} 60.97 & 0.82 & \cellcolor{softgreen} 0.80 & 1.44 & \cellcolor{softgreen} 1.18 & 17.73 & \cellcolor{softgreen} 16.44 & 26.06 & \cellcolor{softgreen} 23.40 \\

\textbf{o4-mini-medium} & \ding{51} & 68.83 & \cellcolor{softgreen} 70.87 & 54.63 & \cellcolor{softgreen} 58.86 & 0.79 & \cellcolor{softgreen} 0.80 & 1.48 & \cellcolor{softgreen} 1.29 & 18.07 & \cellcolor{softgreen} 16.85 & 26.43 & \cellcolor{softgreen} 23.78 \\

\textbf{o4-mini-high} & \ding{51} & 69.56 & \cellcolor{softgreen} 72.89 & 55.72 & \cellcolor{softgreen} 61.33 & 0.79 & \cellcolor{softgreen} 0.46 & 1.51 & \cellcolor{softgreen} 1.34 & 18.03 & \cellcolor{softgreen} 17.50 & 26.14 & \cellcolor{softgreen} 24.51 \\

\textbf{Gemini-2.5} & \ding{51} & \bf 72.06 & \cellcolor{softgreen} \bf 76.34 & 60.67 & \cellcolor{softgreen} 65.95 & 0.80 & \cellcolor{softgreen} 0.79 & \bf 3.15 & \cellcolor{softgreen} \bf  2.61 & \bf 25.36 & \cellcolor{softgreen} \bf 24.21 & \bf 28.61 & \cellcolor{softgreen} 26.15 \\

\textbf{DeepSeek-R1} & \ding{51} & 67.87 & \cellcolor{softgreen} 72.51 & 54.79 & \cellcolor{softgreen} 60.41 & 0.76 & \cellcolor{softgreen} 0.72 & 1.57 & \cellcolor{softgreen} 1.37 & 17.25 & \cellcolor{softgreen} 16.83 & 28.31 & \cellcolor{softgreen} \bf 26.29 \\

\textbf{Claude-3.7} & \ding{55} & 64.57 & \cellcolor{softgreen} 70.04 & 51.56 & \cellcolor{softgreen} 58.60 & 0.76 & \cellcolor{softgreen} 0.68 & 2.50 & \cellcolor{softgreen} 2.20 & 22.17 & \cellcolor{softgreen} 22.81 & 26.42 & \cellcolor{softgreen} 24.90 \\

\textbf{Claude-3.7} & \ding{51} & 71.40 & \cellcolor{softgreen} 74.65 & 58.59 & \cellcolor{softgreen} 60.56 & 0.69 & \cellcolor{softgreen} 0.65 & 2.17 & \cellcolor{softgreen} 2.00 & 21.12 & \cellcolor{softgreen} 21.41 & 26.82 & \cellcolor{softgreen} 25.13 \\

\textbf{Grok-3} & \ding{55} & 66.90 & \cellcolor{softgreen} 72.69 & 52.92 & \cellcolor{softgreen} 59.14 & \bf 0.67 & \cellcolor{softgreen} \bf 0.59 & 2.37 & \cellcolor{softgreen} 1.70 & 24.67 & \cellcolor{softgreen} 24.07 & 26.06 & \cellcolor{softgreen} 23.55 \\

\textbf{Mistral-small} & \ding{55} & 57.48 & \cellcolor{softgreen} 66.03 & 44.34 & \cellcolor{softgreen} 50.92 & 0.75 & \cellcolor{softgreen} 0.65 & 1.84 & \cellcolor{softgreen} 1.56 & 20.37 & \cellcolor{softgreen} 20.62 & 25.58 & \cellcolor{softgreen} 23.87 \\

\textbf{Qwen-3-8b} & \ding{51} & 57.44 & \cellcolor{softgreen} 61.28 & 41.04 & \cellcolor{softgreen} 45.38 & 0.90 & \cellcolor{softgreen} 0.79 & 0.97 & \cellcolor{softgreen} 0.67 & 15.62 & \cellcolor{softgreen} 14.74 & 23.51 & \cellcolor{softgreen} 21.45 \\

\textbf{Qwen-3-30b-a3b} & \ding{51} & 61.57 & \cellcolor{softgreen} 65.19 & 44.42 & \cellcolor{softgreen} 49.83 & 0.76 & \cellcolor{softgreen} 0.70 & 0.92 & \cellcolor{softgreen} 0.81 & 15.39 & \cellcolor{softgreen} 15.59 & 24.49 & \cellcolor{softgreen} 22.16 \\

\textbf{Qwen-3-230b-a22b} & \ding{51} & 62.50 & \cellcolor{softgreen} 68.66 & 49.84 & \cellcolor{softgreen} 55.64 & 0.84 & \cellcolor{softgreen} 0.82 & 0.82 & \cellcolor{softgreen} 0.82 & 14.42 & \cellcolor{softgreen} 14.31 & 22.64 & \cellcolor{softgreen} 21.23 \\

\textbf{Llama-3.1-8b} & \ding{55} & 35.33 & \cellcolor{softgreen} 40.25 & 23.17 & \cellcolor{softgreen} 26.07 & 1.13 & \cellcolor{softgreen} 1.10 & 0.64 & \cellcolor{softgreen} 0.54 & 15.97 & \cellcolor{softgreen} 15.54 & 21.47 & \cellcolor{softgreen} 19.61 \\

\textbf{Llama-3.1-70b} & \ding{55} & 53.28 & \cellcolor{softgreen} 59.12 & 42.46 & \cellcolor{softgreen} 45.04 & 0.79 & \cellcolor{softgreen} 0.73 & 1.24 & \cellcolor{softgreen} 0.83 & 17.37 & \cellcolor{softgreen} 16.38 & 23.42 & \cellcolor{softgreen} 21.25 \\

\textbf{Llama-3.3-70b} & \ding{55} & 56.48 & \cellcolor{softgreen} 63.12 & 41.84 & \cellcolor{softgreen} 46.69 & 0.95 & \cellcolor{softgreen} 0.92 & 1.52 & \cellcolor{softgreen} 1.17 & 19.28 & \cellcolor{softgreen} 19.46 & 21.48 & \cellcolor{softgreen} 20.26 \\

\midrule

\textbf{Average} & - & 63.13 & \cellcolor{softgreen} 67.83 & 50.16 & \cellcolor{softgreen} 54.92 & 0.80 & \cellcolor{softgreen} 0.73 & 1.54 & \cellcolor{softgreen} 1.30 & 18.54 & \cellcolor{softgreen} 18.2 & 24.99 & \cellcolor{softgreen} 23.14 \\



\midrule

\textbf{Human} & - & 79.90 & \cellcolor{softgreen} 76.66 & 60.53 & \cellcolor{softgreen} 62.25 & 0.99 & \cellcolor{softgreen} 0.72 & 5.21 & \cellcolor{softgreen} 5.20 & 25.28 & \cellcolor{softgreen} 28.39 & 33.50 & \cellcolor{softgreen} 31.60 \\

\bottomrule
\end{tabular}}
\caption{Main results of the \texttt{Zero-shot-naive} setting. We color the \colorbox{softgreen}{secondary} results. The result of the \textit{Human} block is collected from English teachers not involved in the dataset construction (Section~\ref{subsec:human_evaluation}). It is not fully comparable to the results of LLMs since human evaluation is conducted on a subset of the full dataset.}
\label{Tab:main_results_zero_shot_naive}
\end{table*}

\paragraph{Evaluation Settings.}
We design three distinct experimental settings to evaluate the LLMs' performance on the \Benchmark{} benchmark under varying levels of instructional detail. This comprehensive approach allows us to systematically assess the LLMs' intrinsic understanding of the task versus their ability to leverage explicit guidance, providing insights into their robustness and adaptability.
(1) \textbf{\texttt{Zero-shot-naive}}: LLMs are provided with only the basic task instruction and the label space for our error taxonomy. Crucially, they do not receive any demonstrations, detailed definitions, or illustrative examples for each error type, nor are they given definitions for the severity scores (1-5). The purpose of this setting is to test the LLMs' unassisted ability to perform fine-grained error analysis with minimal contextual information, thereby establishing a baseline for their inherent capabilities.
(2) \textbf{\texttt{One-shot-detailed}}: The setting builds directly upon the \texttt{Zero-shot-naive} setting by offering more comprehensive guidance and demonstrations. Models receive a detailed definition and example for every error type within our taxonomy, along with an explicit definition for each severity score from 1 to 5. The given demonstration allows us to investigate the LLMs' capacity for in-context learning and their ability to generalize effectively from a specific, relevant example.

The progression from \texttt{Zero-shot-naive} to \texttt{One-shot-detailed} systematically increases the richness and explicitness of the input prompt. This design gradually reduces the task's inherent ambiguity and difficulty for the models.


\subsection{Evaluation Results}
The comprehensive results of our experiments across the three evaluation settings are presented in Table~\ref{Tab:main_results_zero_shot_naive} for \texttt{Zero-shot-naive} and Table~\ref{Tab:main_results_one_shot_detailed} for \texttt{One-shot-detailed}. Our analysis across these settings reveals several key insights into the capabilities and limitations of current LLMs.

\begin{table*}[tb!]
\centering
\renewcommand{\arraystretch}{1.1}
\renewcommand{\tabcolsep}{3pt}
\resizebox{0.89\linewidth}{!}{
\begin{tabular}{lccccccccccccc}
\toprule

\multirow{2}{*}{\textbf{Model}}
& \multirow{2}{*}{\textbf{Think}}


& \multicolumn{4}{c}{\textbf{Classification}}
& \multicolumn{2}{c}{\textbf{Severity}}
& \multicolumn{6}{c}{\textbf{Explanation}}

\\

\cmidrule(ll){3-6} \cmidrule(ll){7-8} \cmidrule(ll){9-14}

&
& \multicolumn{2}{c}{\textbf{Acc}$\uparrow$}
& \multicolumn{2}{c}{\textbf{F$_1$}$\uparrow$}
& \multicolumn{2}{c}{\textbf{MAE}$\downarrow$}
& \multicolumn{2}{c}{\textbf{BLEU}$\uparrow$} 
& \multicolumn{2}{c}{\textbf{METEOR}$\uparrow$}
& \multicolumn{2}{c}{\textbf{ROUGE}$\uparrow$} \\

\midrule

\textbf{GPT-4o} & \ding{55} & 66.57 & \cellcolor{softgreen} 72.02 & 52.97 & \cellcolor{softgreen} 57.66 & 0.78 & \cellcolor{softgreen} 0.66 & 2.46 & \cellcolor{softgreen} 2.16 & 19.96 & \cellcolor{softgreen} 19.70 & 28.41 & \cellcolor{softgreen} 26.13 \\

\textbf{o1} & \ding{51} & 74.49 & \cellcolor{softgreen} 77.80 & 63.94 & \cellcolor{softgreen} 65.57 & 0.82 & \cellcolor{softgreen} 0.69 & 2.27 & \cellcolor{softgreen} 2.61 & 19.04 & \cellcolor{softgreen} 20.49 & 29.42 & \cellcolor{softgreen} \bf 29.11 \\

\textbf{o3-low} & \ding{51} & \bf 76.26 & \cellcolor{softgreen} \bf 78.45 & \bf 65.95 & \cellcolor{softgreen} 65.20 & 0.77 & \cellcolor{softgreen} 0.69 & 2.30 & \cellcolor{softgreen} 2.01 & 19.96 & \cellcolor{softgreen} 19.04 & 27.79 & \cellcolor{softgreen} 25.61 \\

\textbf{o4-mini-low} & \ding{51} & 73.43 & \cellcolor{softgreen} 75.10 & 62.98 & \cellcolor{softgreen} 62.36 & 0.86 & \cellcolor{softgreen} 0.81 & 2.43 & \cellcolor{softgreen} 2.35 & 19.86 & \cellcolor{softgreen} 19.54 & 28.91 & \cellcolor{softgreen} 27.07 \\

\textbf{o4-mini-medium} & \ding{51} & 73.19 & \cellcolor{softgreen} 75.24 & 60.57 & \cellcolor{softgreen} 62.80 & 0.86 & \cellcolor{softgreen} 0.79 & 2.63 & \cellcolor{softgreen} 2.32 & 20.47 & \cellcolor{softgreen} 19.66 & 29.43 & \cellcolor{softgreen} 27.03 \\

\textbf{o4-mini-high} & \ding{51} & 73.53 & \cellcolor{softgreen} 74.79 & 64.71 & \cellcolor{softgreen} 62.04 & 0.86 & \cellcolor{softgreen} 0.80 & 2.58 & \cellcolor{softgreen} 2.28 & 20.22 & \cellcolor{softgreen} 19.77 & 28.99 & \cellcolor{softgreen} 27.28 \\

\textbf{Gemini-2.5} & \ding{51} & 76.19 & \cellcolor{softgreen} 77.17 & 65.60 & \cellcolor{softgreen} 64.79 & 0.75 & \cellcolor{softgreen} 0.74 & \bf 4.29 & \cellcolor{softgreen} \bf 3.57 & \bf 26.76 & \cellcolor{softgreen} \bf 25.42 & \bf 31.36 & \cellcolor{softgreen} 28.06 \\

\textbf{DeepSeek-R1} & \ding{51} & 71.23 & \cellcolor{softgreen} 75.53 & 60.65 & \cellcolor{softgreen} \bf 66.38 & 0.90 & \cellcolor{softgreen} 0.75 & 1.90 & \cellcolor{softgreen} 1.71 & 17.71 & \cellcolor{softgreen} 17.27 & 28.14 & \cellcolor{softgreen} 25.42 \\

\textbf{Claude-3.7} & \ding{55} & 69.50 & \cellcolor{softgreen} 75.73 & 55.38 & \cellcolor{softgreen} 62.69 & \bf 0.71 & \cellcolor{softgreen} \bf 0.63 & 3.61 & \cellcolor{softgreen} 3.24 & 22.67 & \cellcolor{softgreen} 22.85 & 29.71 & \cellcolor{softgreen} 27.74 \\

\textbf{Claude-3.7} & \ding{51} & 73.99 & \cellcolor{softgreen} 77.02 & 59.98 & \cellcolor{softgreen} 63.32 & 0.74 & \cellcolor{softgreen} 0.68 & 3.61 & \cellcolor{softgreen} 3.21 & 22.60 & \cellcolor{softgreen} 22.66 & 29.73 & \cellcolor{softgreen} 27.22 \\

\textbf{Grok-3} & \ding{55} & 67.10 & \cellcolor{softgreen} 74.09 & 52.99 & \cellcolor{softgreen} 60.29 & 0.97 & \cellcolor{softgreen} 0.88 & 3.63 & \cellcolor{softgreen} 2.76 & 24.17 & \cellcolor{softgreen} 23.43 & 29.60 & \cellcolor{softgreen} 26.86 \\

\textbf{Mistral-small} & \ding{55} & 63.64 & \cellcolor{softgreen} 68.68 & 49.69 & \cellcolor{softgreen} 52.93 & 0.94 & \cellcolor{softgreen} 0.81 & 3.06 & \cellcolor{softgreen} 3.33 & 22.14 & \cellcolor{softgreen} 23.53 & 30.50 & \cellcolor{softgreen} 29.26 \\

\textbf{Qwen-230b-a22b} & \ding{51} & 67.40 & \cellcolor{softgreen} 71.75 & 55.33 & \cellcolor{softgreen} 59.30 & 0.85 & \cellcolor{softgreen} 0.73 & 1.85 & \cellcolor{softgreen} 1.64 & 17.68 & \cellcolor{softgreen} 17.62 & 26.26 & \cellcolor{softgreen} 24.45 \\

\textbf{Qwen-30b-a3b} & \ding{51} & 63.47 & \cellcolor{softgreen} 68.71 & 47.22 & \cellcolor{softgreen} 54.33 & 1.05 & \cellcolor{softgreen} 0.84 & 1.33 & \cellcolor{softgreen} 1.34 & 16.69 & \cellcolor{softgreen} 17.25 & 26.05 & \cellcolor{softgreen} 24.57 \\

\textbf{Qwen-8b} & \ding{51} & 57.81 & \cellcolor{softgreen} 62.65 & 41.26 & \cellcolor{softgreen} 47.45 & 0.98 & \cellcolor{softgreen} 0.82 & 1.69 & \cellcolor{softgreen} 1.48 & 17.62 & \cellcolor{softgreen} 17.56 & 26.46 & \cellcolor{softgreen} 24.24 \\

\textbf{Llama-3.1-8b} & \ding{55} & 37.43 & \cellcolor{softgreen} 41.49 & 24.61 & \cellcolor{softgreen} 27.27 & 1.02 & \cellcolor{softgreen} 0.98 & 1.24 & \cellcolor{softgreen} 1.14 & 17.91 & \cellcolor{softgreen} 17.56 & 24.56 & \cellcolor{softgreen} 24.36 \\

\textbf{Llama-3.1-70b} & \ding{55} & 59.51 & \cellcolor{softgreen} 64.30 & 46.10 & \cellcolor{softgreen} 49.95 & 0.81 & \cellcolor{softgreen} 0.72 & 1.55 & \cellcolor{softgreen} 1.41 & 17.04 & \cellcolor{softgreen} 16.27 & 25.72 & \cellcolor{softgreen} 23.38 \\

\textbf{Llama-3.3-70b} & \ding{55} & 61.17 & \cellcolor{softgreen} 65.53 & 47.26 & \cellcolor{softgreen} 49.63 & 0.73 & \cellcolor{softgreen} 0.67 & 2.16 & \cellcolor{softgreen} 2.30 & 19.27 & \cellcolor{softgreen} 20.30 & 26.38 & \cellcolor{softgreen} 24.66 \\

\midrule

\textbf{Average} & - & 66.85 & \cellcolor{softgreen} 70.61 & 54.22 & \cellcolor{softgreen} 57.13 & 0.86 & \cellcolor{softgreen} 0.77 & 2.41 & \cellcolor{softgreen} 2.21 & 19.95 & \cellcolor{softgreen} 19.83 & 28.10 & \cellcolor{softgreen} 26.16 \\

\midrule

\textbf{Human} & - & 79.90 & \cellcolor{softgreen} 76.66 & 60.53 & \cellcolor{softgreen} 62.25 & 0.99 & \cellcolor{softgreen} 0.72 & 5.21 & \cellcolor{softgreen} 5.20 & 25.28 & \cellcolor{softgreen} 28.39 & 33.50 & \cellcolor{softgreen} 31.60 \\

\bottomrule
\end{tabular}}
\caption{Main results of the \texttt{One-shot-detailed} setting. We color the \colorbox{softgreen}{secondary} results. The result of the \textit{Human} block is collected from English teachers not involved in the dataset construction (Section~\ref{subsec:human_evaluation}). It is not fully comparable to the results of LLMs since human evaluation is conducted on a subset of the full dataset.}
\label{Tab:main_results_one_shot_detailed}
\end{table*}

\paragraph{Overview results.}
The overall results reveal that no LLM consistently outperforms others across all three sub-tasks. (1) For \textit{Error Classification}, larger models often designated as thinking models, including Gemini-2.5-pro, o3-low, o1, o4-mini, Claude-3.7-Thinking, and DeepSeek-R1, generally demonstrate superior accuracy. This may be attributed to their enhanced reasoning capabilities, which are beneficial for systematically applying our taxonomy to complex linguistic errors. (2) In the \textit{Error Severity Rating} task under the \texttt{Zero-shot-naive} setting, models such as Grok-3, o1, o3-low, and Claude-3.7-Thinking showed stronger performance, potentially reflecting better intuitive calibration for impact assessment. However, with the provision of detailed definitions and an example in the \texttt{One-shot-detailed} setting, models like Claude-3.7, Llama-3.3-70b, and Claude-3.7-Thinking excelled. (3) For the \textit{Error Explanation} task, all LLMs show low BLEU, METEOR, and ROUGE scores compared to other NLP generation tasks such as machine translation, indicating the difficulty and subjectivity of the task. Specifically, Gemini-2.5-pro exhibited a significant lead, followed by models like Claude-3.7, o1, o3-low, and o4-mini. This highlights that models with strong generative capabilities are better suited for producing high-quality and pedagogically relevant textual feedback.

\paragraph{For the error classification task.}
We observe that classifying errors in essays from elementary school students is generally more challenging than for those from secondary school students. Across various models, the accuracy (Acc) for elementary-level essays is typically 2$\sim$6 percentage points lower than for secondary-level essays. We hypothesize this is primarily because elementary students are more prone to making compound errors, where a single edit may involve multiple intertwined linguistic issues, often spanning more words (Table~\ref{Tab:statistics}). This inherent complexity in the error instances naturally increases the difficulty of assigning a single salient error category. Furthermore, the Macro F1 scores for error classification are consistently and significantly lower than accuracy scores across all models and settings. This discrepancy indicates that while models may perform reasonably well on frequent error types, their ability to accurately classify less frequent error types remains limited. A detailed analysis of model performance on each specific error type is presented in Appendix~\ref{app:extra_results}. 

\paragraph{For the connection between different sub-tasks.}
A notable trend emerging from our results is that models exhibiting superior performance in the Error Classification task also tend to generate higher-quality explanations. It suggests an intrinsic link between the ability to categorize an error and the ability to articulate a meaningful and pedagogically sound explanation for it. This finding highlights the importance of understanding accurate errors as a foundational prerequisite for effective feedback. To further investigate this phenomenon and the potential benefits of optimizing this interplay, we conduct an ablation study, presented in Appendix~\ref{subsec:ablation_prediction_order}, to explore the impact of varying execution order.

\paragraph{For the impact of the information richness of the prompt.}
Our experiments demonstrate a clear positive correlation between the richness of information provided in the prompt and the models' performance in both error classification and explanation. Specifically, transitioning from \texttt{Zero-shot-naive} to \texttt{One-shot-detailed} leads to a marked improvement in average classification Accuracy, Macro F1 scores, and all explanation metrics. It reveals that the problem of fine-grained error analysis can derive substantial benefit from clear definitions and concrete examples. This highlights the differential impact of effective prompt engineering strategies on performance.


\paragraph{For the impact of Thinking Models.}
Comparing the performance of models with explicit ``thinking'' or chain-of-thought-like mechanisms, such as Claude-3.7-Sonnet-Thinking, against their base counterparts (e.g., Claude-3.7-Sonnet), reveals interesting patterns. The Thinking variant consistently achieves significantly higher Accuracy and Macro F1 scores across the different settings in the error classification task. However, their performance on error severity rating and explanation quality remains comparable to the non-thinking versions. This suggests that while structured reasoning processes can enhance the ability to dissect and categorize errors accurately, they may not confer a similar advantage for tasks perceived as more intuitive or requiring nuanced pedagogical capability. This distinction helps isolate the specific benefits of such reasoning mechanisms and points to areas where other approaches might be needed.

\paragraph{For the model performance of different scale parameters.}
In line with general expectations from scaling laws, we observe that larger models typically yield better performance across the tasks in the \Benchmark{} benchmark. For instance, within the Qwen3 series, there is a general trend of improvement as model size increases from Qwen-3-8B to Qwen-3-30B-A3B, and further to Qwen-3-230B-A22B, across all three evaluation settings. This indicates that increased capacity often leads to better generalization and task execution. However, there are exceptions to this trend. For example, Qwen-3-30B-A3B occasionally exhibits slightly superior performance on certain explanation quality metrics compared to the larger Qwen-3-230B-A22B. We attribute this to the specific pre-training and fine-tuning objectives of the Qwen-3 series, which have a strong emphasis on mathematical and coding reasoning tasks. This enhanced reasoning capability, while beneficial for many applications, may not directly or fully translate to the nuanced pedagogical communication and descriptive abilities required by our benchmark. Therefore, advancing model alignment for educational tasks and enhancing their capacity for precise, pedagogically sound error description remain significant research challenges.

\subsection{Human Evaluation Results.}
\label{subsec:human_evaluation}
To gauge the performance gap between LLMs and human intelligence, we engaged several English teachers to conduct fine-grained error analysis on a randomly selected subset of 500 errors from elementary school essays and an additional 500 from secondary school essays. These teachers were not involved in the dataset construction and received no specialized training. This design choice allows us to benchmark LLM performance against the capabilities of human teachers operating without extensive task-specific instruction.

For both error classification and explanation, human teachers almost outperform all evaluated LLMs, particularly under the \texttt{Zero-shot-naive} setting. This underscores a considerable gap between current AI capabilities and human-level expertise in the nuanced task of fine-grained error analysis. The results validate that our benchmark effectively identifies areas where LLMs have substantial room for improvement, thereby justifying its utility in driving research. While enriching prompts with detailed instructions and in-context examples does improve LLM performance and narrow this gap, LLMs still often require more extensive contextual information than humans. Moreover, their generated explanations, even when technically correct, can sometimes lack the conciseness, pedagogical appropriateness, or adaptive nuance of human-authored feedback.
















\definecolor{myblue}{RGB}{32,116,184}
\definecolor{mybgblue}{RGB}{33,114,180}

\begin{figure*}[tb!]
\begin{tcolorbox}[
    colback=mybgblue!20,
    colframe=myblue,
    coltitle=white,
    fonttitle=\bfseries,
    title=Example\label{color_box:case_study},
    overlay={
        \draw[myblue, dashed] 
        (frame.west) -- (frame.east) 
        node[pos=0.5, above, yshift=3mm, font=\small] {PP}
        node[pos=0.5, below, yshift=-3mm, font=\small] {RR};
    }
]
\small
\textbf{Error Sentence}: Mr. Cool Clothes Store's clothes \Red{\underline{is} very \underline{check}} now.
\\
\textbf{Correct Sentence}: Mr. Cool Clothes Store's clothes \Red{\underline{are} very \underline{cheap}} now.
\\ \\
\textbf{Ground Truth}: Error Type: Subject-Verb Agreement Error, Error Severity: 4 \\
Error Explanation: \ding{172} Subject-Verb Agreement Error: The subject 〈clothes〉 is plural and requires the plural verb [are], so 〈is〉 is changed to [are] to conform to grammatical rules; \ding{173} Adjective Error: 〈check〉 means to examine; the adjective [cheap] indicating price should be used here, so 〈check〉 is changed to [cheap] to convey the correct meaning.
\\ \\
\textbf{GPT-4o}: Error Type: \Red{Verb Choice Error}, Error Severity: 4 \\
Error Explanation: \ding{172} The verb 〈is〉 is incorrectly used for the plural subject 〈clothes〉, violating subject-verb agreement rules. The correction [are] ensures proper agreement. \ding{173} The \Red{adjective} 〈check〉 is semantically inappropriate in this context, as it does not convey the intended meaning of being inexpensive. The correction [cheap] provides the proper adjective to convey the intended meaning.
\\ \\
\textbf{o3-low}: Error Type: Subject-Verb Agreement Error, Error Severity: \Red{3} \\
Error Explanation: 〈is〉 violates the rule that the plural subject ``clothes'' requires a plural verb; [are] restores subject-verb agreement. \ding{173}The word 〈check〉 is a misspelling/wrong form of the adjective “cheap”; replacing it with [cheap] provides the correct meaning.
\\ \\
\textbf{Gemini-2.5-pro}: Error Type: \Red{PoS Confusion Error}, Error Severity: 4 \\
Error Explanation: \ding{172} Subject-Verb Agreement Error: The subject of the sentence is ``clothes,'' which is a plural noun. The verb used, 〈is〉, is singular. The verb must agree in number with its subject, so the plural verb [are] is required; \ding{173} \Red{PoS Confusion Error}: The word 〈check〉 is used incorrectly here. The context requires an adjective following ``very'' to describe the clothes. 〈check〉 is typically a verb or a noun. The correct adjective meaning ``low in price'' is [cheap].
\end{tcolorbox}
\caption{A case of error analysis.}
\end{figure*}

\subsection{Case Study}
We present a case study analyzing a complex error from representative LLMs. We notice that GPT-4o misclassifies the error type for the entire edit as ``Verb Choice Error.'' While its explanation correctly identifies the subject-verb agreement violation for "is" $\to$ "are", this is not reflected in the main error type assigned. Secondly, in its explanation for the second error component ("check" $\to$ "cheap"), GPT-4o mistakenly refers to the incorrect word ``check'' as an ``adjective''. On the other hand, the semantic content of o3-mini's explanation is reasonable, but the output does not fully adhere to the desired structured formatting conventions exemplified in the ground truth. For instance, it omits the explicit labeling of each error component within the explanation. Adherence to such formatting is not merely stylistic; it is crucial for consistent automated parsing of results and for providing clear, standardized feedback to learners. Gemini-2.5-pro classifies the ``check'' $\to$ ``cheap'' error as a PoS Confusion Error, which is wrong since they do not have the same root or affix.

This typical case study highlights several recurring challenges for LLMs in our experiments: accurately determining the single most salient and correct error type when multiple errors are present in an edit, consistently applying specific error taxonomies, and adhering strictly to specified output formatting. It underscores that the intricate reasoning and didactic skills essential for granular educational feedback are still developing in LLMs.

\section{Related Work}
\label{sec:related_work}

\paragraph{LLMs for Education.}
Recent advances in LLMs have spurred a wide range of educational applications, including answer grading~\citep{schneider2023towards,chu2025enhancing}, educational question generation~\citep{li-zhang-2024-planning,biancini2024multiple}, interactive educational chatbots~\citep{dan2023educhat,lieb2024student,wang-etal-2024-book2dial}, and classroom simulation~\citep{zhang-etal-2025-simulating,gao2025agent4edu,yue2024mathvc}. These systems leverage deep generative capabilities to provide personalized feedback~\citep{borges-etal-2024-teach,nair-etal-2024-closing,zhang2025sefl} and scaffold learners' understanding~\citep{liu2024socraticlm,scarlatos2025exploring}. For instance, LLM-based tutoring systems have shown promise in delivering real-time corrections and explanations for complex tasks~\citep{treviso-etal-2024-xtower,kim2024reex,yu2024mind}. However, ensuring that LLM-based feedback is accurate, bias-free, and pedagogically grounded remains an open challenge~\citep{ye2025position,chu2025llm,tang2025gmsa}.
Our work fills the gap in the lack of error analysis in the context of language education.

\paragraph{LLMs for Language Learning.}
A growing body of work has specifically examined how LLMs can support second-language (L2) learners~\cite{li2024rethinking,ye-etal-2023-system,ye2022focus}. Early efforts in automated essay scoring leveraged feature engineering or neural networks to produce holistic ratings~\citep{taghipour-ng-2016-neural,ke2019automated,su2025essayjudge}, while more recent studies exploit instruction-tuned LLMs to generate both scores and rubric-based rationales~\citep{chu-etal-2025-rationale,do2025teach}. In grammatical error correction~\citep{ye-etal-2023-mixedit,ye2025corrections,ye2025excgec,qin2025cl,li-etal-2025-rethinking}, LLM prompting has been shown to narrow the gap between supervised systems and human annotators~\citep{li-wang-2024-detection,kong2025scholargec}.
Despite encouraging results, evaluations reveal that LLM-generated feedback may be overly generic or introduce hallucinated corrections~\citep{han-etal-2024-llm,rudian2025feedback,ye-etal-2025-productagent}, underscoring the need for fine-grained and pedagogically sound analysis—a gap our benchmark explicitly targets.


\section{Conclusion}
This paper defines the problem of Fine-grained Error Analysis for English Learners and introduces the \Benchmark{} benchmark. Our extensive evaluation of various LLMs on \Benchmark{} revealed significant challenges. While LLMs demonstrate foundational capabilities, they struggle with consistently applying fine-grained error categories to complex student errors, often lack the pedagogical nuance and conciseness of human feedback, and exhibit performance heavily influenced by prompt engineering, model scale, and internal reasoning structures.



\section*{Limitations}
\label{sec:limitations}

While our study provides the first large-scale benchmark and systematic evaluation for fine-grained error analysis in K-12 English writing, several practical constraints and design choices limit the scope of the current work. We summarize the most salient limitations below.

\paragraph{K-12 focus and domain coverage.}
All source essays are drawn from elementary and secondary school learners. This design serves our educational goal, yet inevitably narrows linguistic variety (e.g., genre complexity, domain-specific vocabulary, discourse structures) compared with adult or professional writing. Consequently, models that perform well on \Benchmark{} are not guaranteed to generalize to university learners, workplace communication, or other L2 populations. Extending the dataset to additional age groups, proficiency levels, and register types is a promising next step.

\paragraph{English-only taxonomy.}
Our error taxonomy is tailored to English morpho-syntax and the curricular requirements of the Chinese K-12 context. Error categories and severity rubrics may not transfer directly to other target languages or educational standards. Multilingual validation and possible language-specific extensions will be required before \Benchmark{} can serve broader data-centric AI research in second-language learning.

\paragraph{Reference-based automatic metrics.}
We evaluate error detection, categorization, and explanation quality primarily with reference-based metrics. Although these metrics allow large-scale reproducible benchmarking, they can over-penalize legitimately different but pedagogically useful feedback, and may not fully capture fluency, readability, or learner uptake. Follow-up work should incorporate rubric-based human ratings or preference learning to complement reference matching.

\section*{Ethics Statement}
\label{sec:ethics_statement}
Every essay in \Benchmark{} was scrubbed of personally identifiable information. We also ensure that no infringement or unethical behavior
occurs during the dataset construction. Experienced teachers involved in the data annotation process were paid \$10 - \$20 per hour, which is well above local minimum wage. To maintain high-quality annotations, we developed a detailed annotation manual and conducted a pre-annotation trial, ensuring that annotators achieved at least 90\% accuracy before the formal annotation process. Any annotator failing to meet this threshold was retrained or replaced. The essay topics and texts are generally concerned with daily life. Therefore, the new research direction and tasks we propose will not cause harm to human society and education applications.

\bibliography{main}

@inproceedings{do-etal-2024-autoregressive-multi,
    title = "Autoregressive Multi-trait Essay Scoring via Reinforcement Learning with Scoring-aware Multiple Rewards",
    author = "Do, Heejin  and
      Ryu, Sangwon  and
      Lee, Gary",
    editor = "Al-Onaizan, Yaser  and
      Bansal, Mohit  and
      Chen, Yun-Nung",
    booktitle = "Proceedings of the 2024 Conference on Empirical Methods in Natural Language Processing",
    month = nov,
    year = "2024",
    address = "Miami, Florida, USA",
    publisher = "Association for Computational Linguistics",
    url = "https://aclanthology.org/2024.emnlp-main.917/",
    doi = "10.18653/v1/2024.emnlp-main.917",
    pages = "16427--16438"
}

@inproceedings{chu-etal-2025-rationale,
    title = "Rationale Behind Essay Scores: Enhancing {S}-{LLM}`s Multi-Trait Essay Scoring with Rationale Generated by {LLM}s",
    author = "Chu, SeongYeub  and
      Kim, Jong Woo  and
      Wong, Bryan  and
      Yi, Mun Yong",
    editor = "Chiruzzo, Luis  and
      Ritter, Alan  and
      Wang, Lu",
    booktitle = "Findings of the Association for Computational Linguistics: NAACL 2025",
    month = apr,
    year = "2025",
    address = "Albuquerque, New Mexico",
    publisher = "Association for Computational Linguistics",
    url = "https://aclanthology.org/2025.findings-naacl.322/",
    pages = "5796--5814",
    ISBN = "979-8-89176-195-7"
}

@article{do2025teach,
  title={Teach-to-Reason with Scoring: Self-Explainable Rationale-Driven Multi-Trait Essay Scoring},
  author={Do, Heejin and Ryu, Sangwon and Lee, Gary Geunbae},
  journal={arXiv preprint arXiv:2502.20748},
  year={2025}
}

@inproceedings{wang-etal-2024-book2dial,
    title = "{B}ook2{D}ial: Generating Teacher Student Interactions from Textbooks for Cost-Effective Development of Educational Chatbots",
    author = "Wang, Junling  and
      Macina, Jakub  and
      Daheim, Nico  and
      Pal Chowdhury, Sankalan  and
      Sachan, Mrinmaya",
    editor = "Ku, Lun-Wei  and
      Martins, Andre  and
      Srikumar, Vivek",
    booktitle = "Findings of the Association for Computational Linguistics: ACL 2024",
    month = aug,
    year = "2024",
    address = "Bangkok, Thailand",
    publisher = "Association for Computational Linguistics",
    url = "https://aclanthology.org/2024.findings-acl.578/",
    doi = "10.18653/v1/2024.findings-acl.578",
    pages = "9707--9731"
}

@inproceedings{li-zhang-2024-planning,
    title = "Planning First, Question Second: An {LLM}-Guided Method for Controllable Question Generation",
    author = "Li, Kunze  and
      Zhang, Yu",
    editor = "Ku, Lun-Wei  and
      Martins, Andre  and
      Srikumar, Vivek",
    booktitle = "Findings of the Association for Computational Linguistics: ACL 2024",
    month = aug,
    year = "2024",
    address = "Bangkok, Thailand",
    publisher = "Association for Computational Linguistics",
    url = "https://aclanthology.org/2024.findings-acl.280/",
    doi = "10.18653/v1/2024.findings-acl.280",
    pages = "4715--4729"
}

@inproceedings{biancini2024multiple,
  title={Multiple-choice question generation using large language models: Methodology and educator insights},
  author={Biancini, Giorgio and Ferrato, Alessio and Limongelli, Carla},
  booktitle={Adjunct Proceedings of the 32nd ACM Conference on User Modeling, Adaptation and Personalization},
  pages={584--590},
  year={2024}
}

@article{dan2023educhat,
  title={Educhat: A large-scale language model-based chatbot system for intelligent education},
  author={Dan, Yuhao and Lei, Zhikai and Gu, Yiyang and Li, Yong and Yin, Jianghao and Lin, Jiaju and Ye, Linhao and Tie, Zhiyan and Zhou, Yougen and Wang, Yilei and others},
  journal={arXiv preprint arXiv:2308.02773},
  year={2023}
}

@inproceedings{lieb2024student,
  title={Student interaction with newtbot: An LLM-as-tutor chatbot for secondary physics education},
  author={Lieb, Anna and Goel, Toshali},
  booktitle={Extended Abstracts of the CHI Conference on Human Factors in Computing Systems},
  pages={1--8},
  year={2024}
}

@inproceedings{zhang-etal-2025-simulating,
    title = "Simulating Classroom Education with {LLM}-Empowered Agents",
    author = "Zhang, Zheyuan  and
      Zhang-Li, Daniel  and
      Yu, Jifan  and
      Gong, Linlu  and
      Zhou, Jinchang  and
      Hao, Zhanxin  and
      Jiang, Jianxiao  and
      Cao, Jie  and
      Liu, Huiqin  and
      Liu, Zhiyuan  and
      Hou, Lei  and
      Li, Juanzi",
    editor = "Chiruzzo, Luis  and
      Ritter, Alan  and
      Wang, Lu",
    booktitle = "Proceedings of the 2025 Conference of the Nations of the Americas Chapter of the Association for Computational Linguistics: Human Language Technologies (Volume 1: Long Papers)",
    month = apr,
    year = "2025",
    address = "Albuquerque, New Mexico",
    publisher = "Association for Computational Linguistics",
    url = "https://aclanthology.org/2025.naacl-long.520/",
    pages = "10364--10379",
    ISBN = "979-8-89176-189-6"
}

@article{gao2025agent4edu,
  title={Agent4Edu: Generating Learner Response Data by Generative Agents for Intelligent Education Systems},
  author={Gao, Weibo and Liu, Qi and Yue, Linan and Yao, Fangzhou and Lv, Rui and Zhang, Zheng and Wang, Hao and Huang, Zhenya},
  journal={arXiv preprint arXiv:2501.10332},
  year={2025}
}

@article{yue2024mathvc,
  title={Mathvc: An llm-simulated multi-character virtual classroom for mathematics education},
  author={Yue, Murong and Lyu, Wenhan and Mifdal, Wijdane and Suh, Jennifer and Zhang, Yixuan and Yao, Ziyu},
  journal={arXiv preprint arXiv:2404.06711},
  year={2024}
}

@article{zhang2025sefl,
  title={SEFL: Harnessing Large Language Model Agents to Improve Educational Feedback Systems},
  author={Zhang, Mike and Dilling, Amalie Pernille and Gondelman, L{\'e}on and Lyngdorf, Niels Erik Ruan and Lindsay, Euan D and Bjerva, Johannes},
  journal={arXiv preprint arXiv:2502.12927},
  year={2025}
}

@inproceedings{borges-etal-2024-teach,
    title = "Let Me Teach You: Pedagogical Foundations of Feedback for Language Models",
    author = {Borges, Beatriz  and
      Tandon, Niket  and
      K{\"a}ser, Tanja  and
      Bosselut, Antoine},
    editor = "Al-Onaizan, Yaser  and
      Bansal, Mohit  and
      Chen, Yun-Nung",
    booktitle = "Proceedings of the 2024 Conference on Empirical Methods in Natural Language Processing",
    month = nov,
    year = "2024",
    address = "Miami, Florida, USA",
    publisher = "Association for Computational Linguistics",
    url = "https://aclanthology.org/2024.emnlp-main.674/",
    doi = "10.18653/v1/2024.emnlp-main.674",
    pages = "12082--12104"
}

@article{yan2024practical,
  title={Practical and ethical challenges of large language models in education: A systematic scoping review},
  author={Yan, Lixiang and Sha, Lele and Zhao, Linxuan and Li, Yuheng and Martinez-Maldonado, Roberto and Chen, Guanliang and Li, Xinyu and Jin, Yueqiao and Ga{\v{s}}evi{\'c}, Dragan},
  journal={British Journal of Educational Technology},
  volume={55},
  number={1},
  pages={90--112},
  year={2024}
}

@inproceedings{nair-etal-2024-closing,
    title = "Closing the Loop: Learning to Generate Writing Feedback via Language Model Simulated Student Revisions",
    author = "Nair, Inderjeet Jayakumar  and
      Tan, Jiaye  and
      Su, Xiaotian  and
      Gere, Anne  and
      Wang, Xu  and
      Wang, Lu",
    editor = "Al-Onaizan, Yaser  and
      Bansal, Mohit  and
      Chen, Yun-Nung",
    booktitle = "Proceedings of the 2024 Conference on Empirical Methods in Natural Language Processing",
    month = nov,
    year = "2024",
    address = "Miami, Florida, USA",
    publisher = "Association for Computational Linguistics",
    url = "https://aclanthology.org/2024.emnlp-main.928/",
    doi = "10.18653/v1/2024.emnlp-main.928",
    pages = "16636--16657"
}

@inproceedings{scarlatos2025exploring,
  title={Exploring knowledge tracing in tutor-student dialogues using llms},
  author={Scarlatos, Alexander and Baker, Ryan S and Lan, Andrew},
  booktitle={Proceedings of the 15th International Learning Analytics and Knowledge Conference},
  pages={249--259},
  year={2025}
}

@article{liu2024socraticlm,
  title={SocraticLM: Exploring socratic personalized teaching with large language models},
  author={Liu, Jiayu and Huang, Zhenya and Xiao, Tong and Sha, Jing and Wu, Jinze and Liu, Qi and Wang, Shijin and Chen, Enhong},
  journal={Advances in Neural Information Processing Systems},
  volume={37},
  pages={85693--85721},
  year={2024}
}

@inproceedings{treviso-etal-2024-xtower,
    title = "x{T}ower: A Multilingual {LLM} for Explaining and Correcting Translation Errors",
    author = "Treviso, Marcos V  and
      Guerreiro, Nuno M  and
      Agrawal, Sweta  and
      Rei, Ricardo  and
      Pombal, Jos{\'e}  and
      Vaz, Tania  and
      Wu, Helena  and
      Silva, Beatriz  and
      Stigt, Daan Van  and
      Martins, Andre",
    editor = "Al-Onaizan, Yaser  and
      Bansal, Mohit  and
      Chen, Yun-Nung",
    booktitle = "Findings of the Association for Computational Linguistics: EMNLP 2024",
    month = nov,
    year = "2024",
    address = "Miami, Florida, USA",
    publisher = "Association for Computational Linguistics",
    url = "https://aclanthology.org/2024.findings-emnlp.892/",
    doi = "10.18653/v1/2024.findings-emnlp.892",
    pages = "15222--15239"
}

@inproceedings{
kim2024reex,
title={Re-Ex: Revising after Explanation reduces the Factual Errors in {LLM} Responses},
author={Juyeon Kim and Jeongeun Lee and YoonHo Chang and CHANYEOL CHOI and Jun-Seong Kim and Jy-yong Sohn},
booktitle={ICLR 2024 Workshop on Reliable and Responsible Foundation Models},
year={2024},
url={https://openreview.net/forum?id=tyEWrLVU1b}
}

@inproceedings{ye2025excgec,
  title={EXCGEC: A Benchmark for Edit-Wise Explainable Chinese Grammatical Error Correction},
  author={Ye, Jingheng and Qin, Shang and Li, Yinghui and Cheng, Xuxin and Qin, Libo and Zheng, Hai-Tao and Shen, Ying and Xing, Peng and Xu, Zishan and Cheng, Guo and others},
  booktitle={Proceedings of the AAAI Conference on Artificial Intelligence},
  volume={39},
  number={24},
  pages={25678--25686},
  year={2025}
}

@article{ye2025position,
  title={Position: LLMs Can be Good Tutors in Foreign Language Education},
  author={Ye, Jingheng and Wang, Shen and Zou, Deqing and Yan, Yibo and Wang, Kun and Zheng, Hai-Tao and Xu, Zenglin and King, Irwin and Yu, Philip S and Wen, Qingsong},
  journal={arXiv preprint arXiv:2502.05467},
  year={2025}
}

@article{chu2025llm,
  title={Llm agents for education: Advances and applications},
  author={Chu, Zhendong and Wang, Shen and Xie, Jian and Zhu, Tinghui and Yan, Yibo and Ye, Jingheng and Zhong, Aoxiao and Hu, Xuming and Liang, Jing and Yu, Philip S and others},
  journal={arXiv preprint arXiv:2503.11733},
  year={2025}
}

@article{chu2025enhancing,
  title={Enhancing LLM-Based Short Answer Grading with Retrieval-Augmented Generation},
  author={Chu, Yucheng and He, Peng and Li, Hang and Han, Haoyu and Yang, Kaiqi and Xue, Yu and Li, Tingting and Krajcik, Joseph and Tang, Jiliang},
  journal={arXiv preprint arXiv:2504.05276},
  year={2025}
}

@article{schneider2023towards,
  title={Towards llm-based autograding for short textual answers},
  author={Schneider, Johannes and Schenk, Bernd and Niklaus, Christina},
  journal={arXiv preprint arXiv:2309.11508},
  year={2023}
}

@inproceedings{ke2019automated,
  title={Automated Essay Scoring: A Survey of the State of the Art.},
  author={Ke, Zixuan and Ng, Vincent},
  booktitle={IJCAI},
  volume={19},
  pages={6300--6308},
  year={2019}
}

@inproceedings{taghipour-ng-2016-neural,
    title = "A Neural Approach to Automated Essay Scoring",
    author = "Taghipour, Kaveh  and
      Ng, Hwee Tou",
    editor = "Su, Jian  and
      Duh, Kevin  and
      Carreras, Xavier",
    booktitle = "Proceedings of the 2016 Conference on Empirical Methods in Natural Language Processing",
    month = nov,
    year = "2016",
    address = "Austin, Texas",
    publisher = "Association for Computational Linguistics",
    url = "https://aclanthology.org/D16-1193/",
    doi = "10.18653/v1/D16-1193",
    pages = "1882--1891"
}

@inproceedings{kong2025scholargec,
  title={ScholarGEC: Enhancing Controllability of Large Language Model for Chinese Academic Grammatical Error Correction},
  author={Kong, Zixiao and Wang, Xianquan and Shen, Shuanghong and Zhu, Keyu and Xu, Huibo and Su, Yu},
  booktitle={Proceedings of the AAAI Conference on Artificial Intelligence},
  volume={39},
  number={23},
  pages={24339--24347},
  year={2025}
}

@inproceedings{li-wang-2024-detection,
    title = "Detection-Correction Structure via General Language Model for Grammatical Error Correction",
    author = "Li, Wei  and
      Wang, Houfeng",
    editor = "Ku, Lun-Wei  and
      Martins, Andre  and
      Srikumar, Vivek",
    booktitle = "Proceedings of the 62nd Annual Meeting of the Association for Computational Linguistics (Volume 1: Long Papers)",
    month = aug,
    year = "2024",
    address = "Bangkok, Thailand",
    publisher = "Association for Computational Linguistics",
    url = "https://aclanthology.org/2024.acl-long.96/",
    doi = "10.18653/v1/2024.acl-long.96",
    pages = "1748--1763"
}

@inproceedings{ye-etal-2023-mixedit,
    title = "{M}ix{E}dit: Revisiting Data Augmentation and Beyond for Grammatical Error Correction",
    author = "Ye, Jingheng  and
      Li, Yinghui  and
      Li, Yangning  and
      Zheng, Hai-Tao",
    editor = "Bouamor, Houda  and
      Pino, Juan  and
      Bali, Kalika",
    booktitle = "Findings of the Association for Computational Linguistics: EMNLP 2023",
    month = dec,
    year = "2023",
    address = "Singapore",
    publisher = "Association for Computational Linguistics",
    url = "https://aclanthology.org/2023.findings-emnlp.681/",
    doi = "10.18653/v1/2023.findings-emnlp.681",
    pages = "10161--10175"
}

@inproceedings{rudian2025feedback,
  title={Feedback on Feedback: Student’s Perceptions for Feedback from Teachers and Few-Shot LLMs},
  author={R{\"u}dian, Sylvio and Podelo, Julia and Ku{\v{z}}{\'\i}lek, Jakub and Pinkwart, Niels},
  booktitle={Proceedings of the 15th International Learning Analytics and Knowledge Conference},
  pages={82--92},
  year={2025}
}

@inproceedings{han-etal-2024-llm,
    title = "{LLM}-as-a-tutor in {EFL} Writing Education: Focusing on Evaluation of Student-{LLM} Interaction",
    author = "Han, Jieun  and
      Yoo, Haneul  and
      Myung, Junho  and
      Kim, Minsun  and
      Lim, Hyunseung  and
      Kim, Yoonsu  and
      Lee, Tak Yeon  and
      Hong, Hwajung  and
      Kim, Juho  and
      Ahn, So-Yeon  and
      Oh, Alice",
    editor = "Kumar, Sachin  and
      Balachandran, Vidhisha  and
      Park, Chan Young  and
      Shi, Weijia  and
      Hayati, Shirley Anugrah  and
      Tsvetkov, Yulia  and
      Smith, Noah  and
      Hajishirzi, Hannaneh  and
      Kang, Dongyeop  and
      Jurgens, David",
    booktitle = "Proceedings of the 1st Workshop on Customizable NLP: Progress and Challenges in Customizing NLP for a Domain, Application, Group, or Individual (CustomNLP4U)",
    month = nov,
    year = "2024",
    address = "Miami, Florida, USA",
    publisher = "Association for Computational Linguistics",
    url = "https://aclanthology.org/2024.customnlp4u-1.21/",
    doi = "10.18653/v1/2024.customnlp4u-1.21",
    pages = "284--293"
}

@article{wang2024large,
  title={Large language models for education: A survey and outlook},
  author={Wang, Shen and Xu, Tianlong and Li, Hang and Zhang, Chaoli and Liang, Joleen and Tang, Jiliang and Yu, Philip S and Wen, Qingsong},
  journal={arXiv preprint arXiv:2403.18105},
  year={2024}
}

@article{xu2024large,
  title={Large language models for education: A survey},
  author={Xu, Hanyi and Gan, Wensheng and Qi, Zhenlian and Wu, Jiayang and Yu, Philip S},
  journal={arXiv preprint arXiv:2405.13001},
  year={2024}
}

@misc{he2025survey,
  title={A survey on complex reasoning of large language models through the lens of self-evolution},
  author={He, Tao and Li, Hao and Chen, Jingchang and Liu, Runxuan and Cao, Yixin and Liao, Lizi and Zheng, Zihao and Chu, Zheng and Liang, Jiafeng and Liu, Ming and others},
  year={2025},
  publisher={February}
}

@article{liu2025erevise+,
  title={eRevise+ RF: A Writing Evaluation System for Assessing Student Essay Revisions and Providing Formative Feedback},
  author={Liu, Zhexiong and Litman, Diane and Wang, Elaine and Li, Tianwen and Gobat, Mason and Matsumura, Lindsay Clare and Correnti, Richard},
  journal={arXiv preprint arXiv:2501.00715},
  year={2025}
}

@inproceedings{stahl-etal-2024-exploring,
    title = "Exploring {LLM} Prompting Strategies for Joint Essay Scoring and Feedback Generation",
    author = "Stahl, Maja  and
      Biermann, Leon  and
      Nehring, Andreas  and
      Wachsmuth, Henning",
    editor = {Kochmar, Ekaterina  and
      Bexte, Marie  and
      Burstein, Jill  and
      Horbach, Andrea  and
      Laarmann-Quante, Ronja  and
      Tack, Ana{\"i}s  and
      Yaneva, Victoria  and
      Yuan, Zheng},
    booktitle = "Proceedings of the 19th Workshop on Innovative Use of NLP for Building Educational Applications (BEA 2024)",
    month = jun,
    year = "2024",
    address = "Mexico City, Mexico",
    publisher = "Association for Computational Linguistics",
    url = "https://aclanthology.org/2024.bea-1.23/",
    pages = "283--298"
}

@article{han2023fabric,
  title={Fabric: Automated scoring and feedback generation for essays},
  author={Han, Jieun and Yoo, Haneul and Myung, Junho and Kim, Minsun and Lim, Hyunseung and Kim, Yoonsu and Lee, Tak Yeon and Hong, Hwajung and Kim, Juho and Ahn, So-Yeon and others},
  journal={arXiv preprint arXiv:2310.05191},
  year={2023}
}

@inproceedings{huang-etal-2023-frustratingly,
    title = "A Frustratingly Easy Plug-and-Play Detection-and-Reasoning Module for {C}hinese Spelling Check",
    author = "Huang, Haojing  and
      Ye, Jingheng  and
      Zhou, Qingyu  and
      Li, Yinghui  and
      Li, Yangning  and
      Zhou, Feng  and
      Zheng, Hai-Tao",
    editor = "Bouamor, Houda  and
      Pino, Juan  and
      Bali, Kalika",
    booktitle = "Findings of the Association for Computational Linguistics: EMNLP 2023",
    month = dec,
    year = "2023",
    address = "Singapore",
    publisher = "Association for Computational Linguistics",
    url = "https://aclanthology.org/2023.findings-emnlp.771/",
    doi = "10.18653/v1/2023.findings-emnlp.771",
    pages = "11514--11525"
}

@article{bryant2023grammatical,
  title={Grammatical error correction: A survey of the state of the art},
  author={Bryant, Christopher and Yuan, Zheng and Qorib, Muhammad Reza and Cao, Hannan and Ng, Hwee Tou and Briscoe, Ted},
  journal={Computational Linguistics},
  volume={49},
  number={3},
  pages={643--701},
  year={2023},
  publisher={MIT Press One Broadway, 12th Floor, Cambridge, Massachusetts 02142, USA~…}
}

@inproceedings{ye-etal-2023-cleme,
    title = "{CLEME}: Debiasing Multi-reference Evaluation for Grammatical Error Correction",
    author = "Ye, Jingheng  and
      Li, Yinghui  and
      Zhou, Qingyu  and
      Li, Yangning  and
      Ma, Shirong  and
      Zheng, Hai-Tao  and
      Shen, Ying",
    editor = "Bouamor, Houda  and
      Pino, Juan  and
      Bali, Kalika",
    booktitle = "Proceedings of the 2023 Conference on Empirical Methods in Natural Language Processing",
    month = dec,
    year = "2023",
    address = "Singapore",
    publisher = "Association for Computational Linguistics",
    url = "https://aclanthology.org/2023.emnlp-main.378/",
    doi = "10.18653/v1/2023.emnlp-main.378",
    pages = "6174--6189"
}

@article{zou2025revisiting,
  title={Revisiting Classification Taxonomy for Grammatical Errors},
  author={Zou, Deqing and Ye, Jingheng and Liu, Yulu and Wu, Yu and Xu, Zishan and Li, Yinghui and Zheng, Hai-Tao and An, Bingxu and Wei, Zhao and Xu, Yong},
  journal={arXiv preprint arXiv:2502.11890},
  year={2025}
}

@book{james2013errors,
  title={Errors in language learning and use: Exploring error analysis},
  author={James, Carl},
  year={2013},
  publisher={Routledge}
}

@article{erdougan2005contribution,
  title={Contribution of error analysis to foreign language teaching},
  author={Erdo{\u{g}}an, Vacide},
  journal={Mersin {\"U}niversitesi E{\u{g}}itim Fak{\"u}ltesi Dergisi},
  volume={1},
  number={2},
  year={2005},
  publisher={Mersin University}
}

@inproceedings{daheim-etal-2024-stepwise,
    title = "Stepwise Verification and Remediation of Student Reasoning Errors with Large Language Model Tutors",
    author = "Daheim, Nico  and
      Macina, Jakub  and
      Kapur, Manu  and
      Gurevych, Iryna  and
      Sachan, Mrinmaya",
    editor = "Al-Onaizan, Yaser  and
      Bansal, Mohit  and
      Chen, Yun-Nung",
    booktitle = "Proceedings of the 2024 Conference on Empirical Methods in Natural Language Processing",
    month = nov,
    year = "2024",
    address = "Miami, Florida, USA",
    publisher = "Association for Computational Linguistics",
    url = "https://aclanthology.org/2024.emnlp-main.478/",
    doi = "10.18653/v1/2024.emnlp-main.478",
    pages = "8386--8411"
}

@article{hurst2024gpt,
  title={Gpt-4o system card},
  author={Hurst, Aaron and Lerer, Adam and Goucher, Adam P and Perelman, Adam and Ramesh, Aditya and Clark, Aidan and Ostrow, AJ and Welihinda, Akila and Hayes, Alan and Radford, Alec and others},
  journal={arXiv preprint arXiv:2410.21276},
  year={2024}
}

@article{jaech2024openai,
  title={Openai o1 system card},
  author={Jaech, Aaron and Kalai, Adam and Lerer, Adam and Richardson, Adam and El-Kishky, Ahmed and Low, Aiden and Helyar, Alec and Madry, Aleksander and Beutel, Alex and Carney, Alex and others},
  journal={arXiv preprint arXiv:2412.16720},
  year={2024}
}

@misc{grok,
  author = {grok},
  title = {Grok 3 Beta — The Age of Reasoning Agents},
  howpublished = {\url{https://x.ai/news/grok-3}},
  year = {2025}
}

@misc{gpto4,
  author = {OpenAI},
  title = {Introducing OpenAI o3 and o4-mini},
  howpublished = {\url{https://openai.com/index/introducing-o3-and-o4-mini/}},
  year = {2025}
}

@article{team2023gemini,
  title={Gemini: a family of highly capable multimodal models},
  author={Team, Gemini and Anil, Rohan and Borgeaud, Sebastian and Alayrac, Jean-Baptiste and Yu, Jiahui and Soricut, Radu and Schalkwyk, Johan and Dai, Andrew M and Hauth, Anja and Millican, Katie and others},
  journal={arXiv preprint arXiv:2312.11805},
  year={2023}
}

@misc{Claude37,
  author = {claude},
  title = {Claude 3.7 Sonnet and Claude Code},
  howpublished = {\url{https://www.anthropic.com/news/claude-3-7-sonnet}},
  year = {2025}
}

@article{guo2025deepseek,
  title={Deepseek-r1: Incentivizing reasoning capability in llms via reinforcement learning},
  author={Guo, Daya and Yang, Dejian and Zhang, Haowei and Song, Junxiao and Zhang, Ruoyu and Xu, Runxin and Zhu, Qihao and Ma, Shirong and Wang, Peiyi and Bi, Xiao and others},
  journal={arXiv preprint arXiv:2501.12948},
  year={2025}
}

@misc{qwen3,
  author = {Qwen Team},
  title = {Qwen3: Think Deeper, Act Faster},
  howpublished = {\url{https://qwenlm.github.io/blog/qwen3/}},
  year = {2025}
}

@article{grattafiori2024llama,
  title={The llama 3 herd of models},
  author={Grattafiori, Aaron and Dubey, Abhimanyu and Jauhri, Abhinav and Pandey, Abhinav and Kadian, Abhishek and Al-Dahle, Ahmad and Letman, Aiesha and Mathur, Akhil and Schelten, Alan and Vaughan, Alex and others},
  journal={arXiv preprint arXiv:2407.21783},
  year={2024}
}

@misc{mistral-small,
  author = {Mistral AI},
  title = {Mistral Small 3.1},
  howpublished = {\url{https://mistral.ai/news/mistral-small-3-1}},
  year = {2025}
}

@inproceedings{zeng-etal-2024-evaluating,
    title = "Evaluating Prompting Strategies for Grammatical Error Correction Based on Language Proficiency",
    author = "Zeng, Min  and
      Kuang, Jiexin  and
      Qiu, Mengyang  and
      Song, Jayoung  and
      Park, Jungyeul",
    editor = "Calzolari, Nicoletta  and
      Kan, Min-Yen  and
      Hoste, Veronique  and
      Lenci, Alessandro  and
      Sakti, Sakriani  and
      Xue, Nianwen",
    booktitle = "Proceedings of the 2024 Joint International Conference on Computational Linguistics, Language Resources and Evaluation (LREC-COLING 2024)",
    month = may,
    year = "2024",
    address = "Torino, Italia",
    publisher = "ELRA and ICCL",
    url = "https://aclanthology.org/2024.lrec-main.569/",
    pages = "6426--6430"
}

@article{ye2024cleme2,
  title={CLEME2. 0: Towards More Interpretable Evaluation by Disentangling Edits for Grammatical Error Correction},
  author={Ye, Jingheng and Xu, Zishan and Li, Yinghui and Cheng, Xuxin and Song, Linlin and Zhou, Qingyu and Zheng, Hai-Tao and Shen, Ying and Su, Xin},
  journal={arXiv preprint arXiv:2407.00934},
  year={2024}
}

@inproceedings{papineni-etal-2002-bleu,
    title = "{B}leu: a Method for Automatic Evaluation of Machine Translation",
    author = "Papineni, Kishore  and
      Roukos, Salim  and
      Ward, Todd  and
      Zhu, Wei-Jing",
    editor = "Isabelle, Pierre  and
      Charniak, Eugene  and
      Lin, Dekang",
    booktitle = "Proceedings of the 40th Annual Meeting of the Association for Computational Linguistics",
    month = jul,
    year = "2002",
    address = "Philadelphia, Pennsylvania, USA",
    publisher = "Association for Computational Linguistics",
    url = "https://aclanthology.org/P02-1040/",
    doi = "10.3115/1073083.1073135",
    pages = "311--318"
}

@inproceedings{banerjee-lavie-2005-meteor,
    title = "{METEOR}: An Automatic Metric for {MT} Evaluation with Improved Correlation with Human Judgments",
    author = "Banerjee, Satanjeev  and
      Lavie, Alon",
    editor = "Goldstein, Jade  and
      Lavie, Alon  and
      Lin, Chin-Yew  and
      Voss, Clare",
    booktitle = "Proceedings of the {ACL} Workshop on Intrinsic and Extrinsic Evaluation Measures for Machine Translation and/or Summarization",
    month = jun,
    year = "2005",
    address = "Ann Arbor, Michigan",
    publisher = "Association for Computational Linguistics",
    url = "https://aclanthology.org/W05-0909/",
    pages = "65--72"
}

@inproceedings{lin2004rouge,
  title={Rouge: A package for automatic evaluation of summaries},
  author={Lin, Chin-Yew},
  booktitle={Text summarization branches out},
  pages={74--81},
  year={2004}
}

@article{yan2024errorradar,
  title={Errorradar: Benchmarking complex mathematical reasoning of multimodal large language models via error detection},
  author={Yan, Yibo and Wang, Shen and Huo, Jiahao and Li, Hang and Li, Boyan and Su, Jiamin and Gao, Xiong and Zhang, Yi-Fan and Xu, Tianlong and Chu, Zhendong and others},
  journal={arXiv preprint arXiv:2410.04509},
  year={2024}
}

@inproceedings{wang-etal-2024-bridging,
    title = "Bridging the Novice-Expert Gap via Models of Decision-Making: A Case Study on Remediating Math Mistakes",
    author = "Wang, Rose  and
      Zhang, Qingyang  and
      Robinson, Carly  and
      Loeb, Susanna  and
      Demszky, Dorottya",
    editor = "Duh, Kevin  and
      Gomez, Helena  and
      Bethard, Steven",
    booktitle = "Proceedings of the 2024 Conference of the North American Chapter of the Association for Computational Linguistics: Human Language Technologies (Volume 1: Long Papers)",
    month = jun,
    year = "2024",
    address = "Mexico City, Mexico",
    publisher = "Association for Computational Linguistics",
    url = "https://aclanthology.org/2024.naacl-long.120/",
    doi = "10.18653/v1/2024.naacl-long.120",
    pages = "2174--2199"
}

@article{su2025essayjudge,
  title={Essayjudge: A multi-granular benchmark for assessing automated essay scoring capabilities of multimodal large language models},
  author={Su, Jiamin and Yan, Yibo and Fu, Fangteng and Zhang, Han and Ye, Jingheng and Liu, Xiang and Huo, Jiahao and Zhou, Huiyu and Hu, Xuming},
  journal={arXiv preprint arXiv:2502.11916},
  year={2025}
}

@article{qin2025cl,
  title={CL $^{2}$ GEC: A Multi-Discipline Benchmark for Continual Learning in Chinese Literature Grammatical Error Correction},
  author={Qin, Shang and Ye, Jingheng and Li, Yinghui and Zheng, Hai-Tao and Li, Qi and Shan, Jinxiao and Li, Zhixing and Kim, Hong-Gee},
  journal={arXiv preprint arXiv:2509.13672},
  year={2025}
}

@article{ye2025corrections,
  title={Corrections Meet Explanations: A Unified Framework for Explainable Grammatical Error Correction},
  author={Ye, Jingheng and Qin, Shang and Li, Yinghui and Zheng, Hai-Tao and Wang, Shen and Wen, Qingsong},
  journal={arXiv preprint arXiv:2502.15261},
  year={2025}
}

@article{ye2022focus,
  title={Focus is what you need for chinese grammatical error correction},
  author={Ye, Jingheng and Li, Yinghui and Ma, Shirong and Xie, Rui and Wu, Wei and Zheng, Hai-Tao},
  journal={arXiv preprint arXiv:2210.12692},
  year={2022}
}

@article{yu2024mind,
  title={Mind scramble: Unveiling large language model psychology via typoglycemia},
  author={Yu, Miao and Mao, Junyuan and Zhang, Guibin and Ye, Jingheng and Fang, Junfeng and Zhong, Aoxiao and Liu, Yang and Liang, Yuxuan and Wang, Kun and Wen, Qingsong},
  journal={arXiv preprint arXiv:2410.01677},
  year={2024}
}

@article{tang2025gmsa,
  title={GMSA: Enhancing Context Compression via Group Merging and Layer Semantic Alignment},
  author={Tang, Jiwei and Zhang, Zhicheng and Wu, Shunlong and Ye, Jingheng and Bai, Lichen and Wang, Zitai and Lu, Tingwei and Chen, Jiaqi and Hai, Lin and Zheng, Hai-Tao and others},
  journal={arXiv preprint arXiv:2505.12215},
  year={2025}
}

@inproceedings{ye-etal-2023-system,
    title = "System Report for {CCL}23-Eval Task 7: {THU} {KEL}ab (sz) - Exploring Data Augmentation and Denoising for {C}hinese Grammatical Error Correction",
    author = "Ye, Jingheng  and
      Li, Yinghui  and
      Zheng, Haitao",
    editor = "Sun, Maosong  and
      Qin, Bing  and
      Qiu, Xipeng  and
      Jiang, Jing  and
      Han, Xianpei",
    booktitle = "Proceedings of the 22nd Chinese National Conference on Computational Linguistics (Volume 3: Evaluations)",
    month = aug,
    year = "2023",
    address = "Harbin, China",
    publisher = "Chinese Information Processing Society of China",
    url = "https://aclanthology.org/2023.ccl-3.29",
    pages = "262--270",
    language = "English",
}

@article{li2024rethinking,
  title={Rethinking the Roles of Large Language Models in Chinese Grammatical Error Correction},
  author={Li, Yinghui and Qin, Shang and Ye, Jingheng and Ma, Shirong and Li, Yangning and Qin, Libo and Hu, Xuming and Jiang, Wenhao and Zheng, Hai-Tao and Yu, Philip S},
  journal={arXiv preprint arXiv:2402.11420},
  year={2024}
}

@inproceedings{ye-etal-2025-productagent,
    title = "{P}roduct{A}gent: Benchmarking Conversational Product Search Agent with Asking Clarification Questions",
    author = "Ye, Jingheng  and
      Jiang, Yong  and
      Wang, Xiaobin  and
      Li, Yinghui  and
      Li, Yangning  and
      Xie, Pengjun  and
      Huang, Fei",
    editor = "Potdar, Saloni  and
      Rojas-Barahona, Lina  and
      Montella, Sebastien",
    booktitle = "Proceedings of the 2025 Conference on Empirical Methods in Natural Language Processing: Industry Track",
    month = nov,
    year = "2025",
    address = "Suzhou (China)",
    publisher = "Association for Computational Linguistics",
    url = "https://aclanthology.org/2025.emnlp-industry.25/",
    doi = "10.18653/v1/2025.emnlp-industry.25",
    pages = "383--398",
    ISBN = "979-8-89176-333-3"
}

@inproceedings{li-etal-2025-rethinking,
    title = "Rethinking the Roles of Large Language Models in {C}hinese Grammatical Error Correction",
    author = "Li, Yinghui  and
      Qin, Shang  and
      Ye, Jingheng  and
      Huang, Haojing  and
      Li, Yangning  and
      Guo, Shu-Yu  and
      Qin, Libo  and
      Hu, Xuming  and
      Jiang, Wenhao  and
      Zheng, Hai-Tao  and
      Yu, Philip S.",
    editor = "Rehm, Georg  and
      Li, Yunyao",
    booktitle = "Proceedings of the 63rd Annual Meeting of the Association for Computational Linguistics (Volume 6: Industry Track)",
    month = jul,
    year = "2025",
    address = "Vienna, Austria",
    publisher = "Association for Computational Linguistics",
    url = "https://aclanthology.org/2025.acl-industry.39/",
    doi = "10.18653/v1/2025.acl-industry.39",
    pages = "553--567",
    ISBN = "979-8-89176-288-6"
}

\appendix

\section{Details of Dataset Construction}

\tikzset{
    root/.style = {align=center, text width=2cm, rounded corners=3pt, line width=0.3mm, fill=gray!10, draw=gray!80, font=\small},
    single_word/.style = {align=center, text width=4cm, rounded corners=3pt, line width=0.3mm, fill=blue!10, draw=blue!80, font=\footnotesize},
    single_word_type/.style = {align=left, text width=10cm, rounded corners=3pt, line width=0.3mm, fill=blue!10, draw=blue!80, font=\footnotesize},
    inter_word/.style = {align=center, text width=4cm, rounded corners=3pt, line width=0.3mm, fill=red!10, draw=red!80, font=\footnotesize},
    inter_word_type/.style = {align=left, text width=10cm, rounded corners=3pt, line width=0.3mm, fill=red!10, draw=red!80, font=\footnotesize},
    discourse/.style = {align=center, text width=4cm, rounded corners=3pt, line width=0.3mm, fill=cyan!10, draw=cyan!80, font=\footnotesize},
    discourse_type/.style = {align=left, text width=10cm, rounded corners=3pt, line width=0.3mm, fill=cyan!10, draw=cyan!80, font=\footnotesize},
}


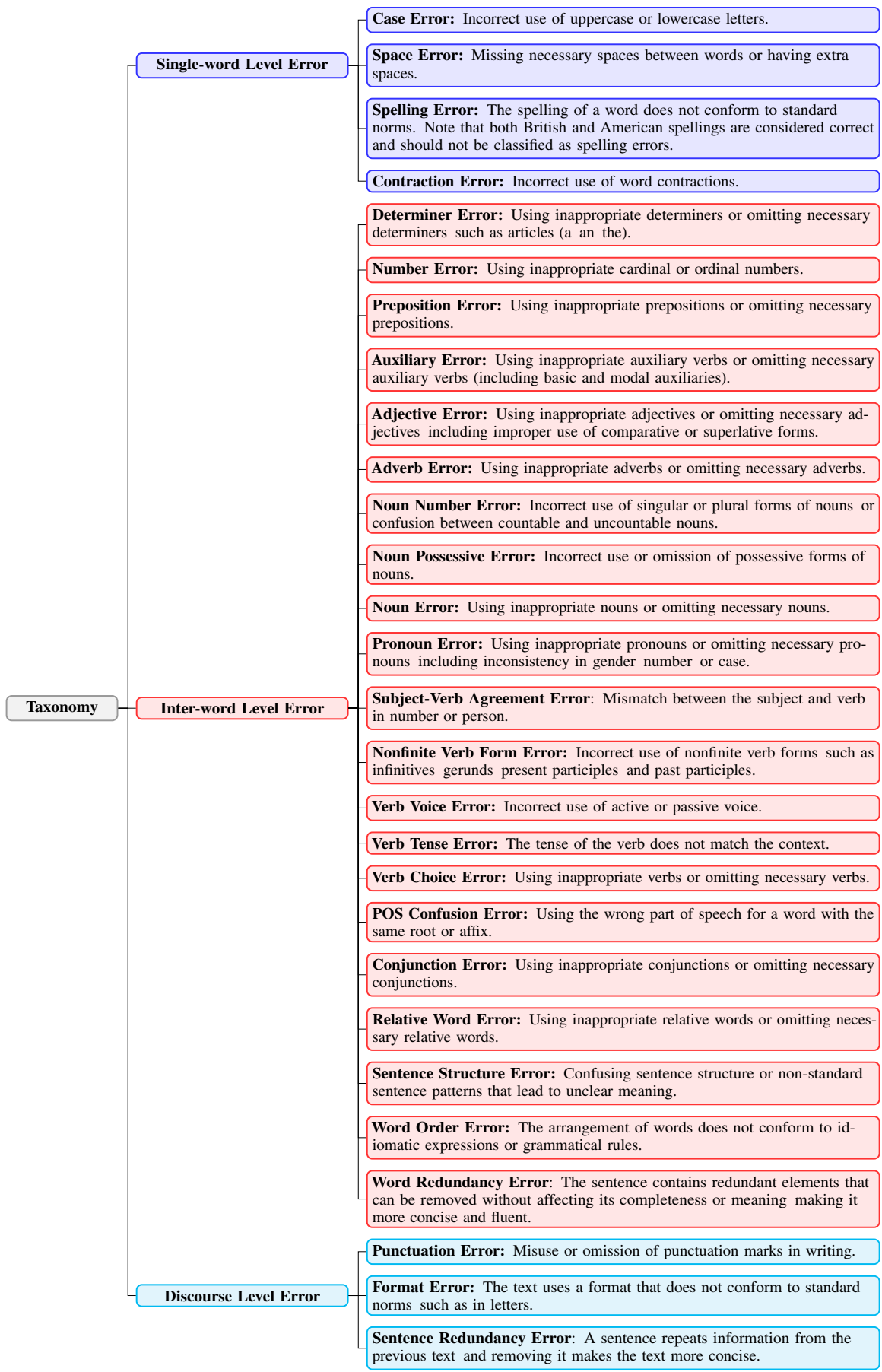
\begin{figure*}
\centering
\small
\resizebox{0.95\linewidth}{!}{
\begin{forest}
    for tree={
        forked edges,
        grow'=0,
        calign=child edge,
        calign child=(n_children()+1)/2,
    },
    [\textbf{Taxonomy}, root
        [\textbf{Single-word Level Error}, single_word
            [\textbf{Case Error:} Incorrect use of uppercase or lowercase letters., single_word_type]
            [\textbf{Space Error:} Missing necessary spaces between words or having extra spaces., single_word_type]
            [\textbf{Spelling Error:} The spelling of a word does not conform to standard norms. Note that both British and American spellings are considered correct and should not be classified as spelling errors., single_word_type]
            [\textbf{Contraction Error:} Incorrect use of word contractions., single_word_type]
        ]
        [\textbf{Inter-word Level Error}, inter_word
            [\textbf{Determiner Error:} Using inappropriate determiners or omitting necessary determiners\, such as articles (a\, an\, the)., inter_word_type]
            [\textbf{Number Error:} Using inappropriate cardinal or ordinal numbers., inter_word_type]
            [\textbf{Preposition Error:} Using inappropriate prepositions or omitting necessary prepositions., inter_word_type]
            [\textbf{Auxiliary Error:} Using inappropriate auxiliary verbs or omitting necessary auxiliary verbs (including basic and modal auxiliaries)., inter_word_type]
            [\textbf{Adjective Error:} Using inappropriate adjectives or omitting necessary adjectives\, including improper use of comparative or superlative forms., inter_word_type]
            [\textbf{Adverb Error:} Using inappropriate adverbs or omitting necessary adverbs., inter_word_type]
            [\textbf{Noun Number Error:} Incorrect use of singular or plural forms of nouns\, or confusion between countable and uncountable nouns., inter_word_type]
            [\textbf{Noun Possessive Error:} Incorrect use or omission of possessive forms of nouns., inter_word_type]
            [\textbf{Noun Error:} Using inappropriate nouns or omitting necessary nouns., inter_word_type]
            [\textbf{Pronoun Error:} Using inappropriate pronouns or omitting necessary pronouns\, including inconsistency in gender\, number\, or case., inter_word_type]
            [\textbf{Subject-Verb Agreement Error}: Mismatch between the subject and verb in number or person., inter_word_type]
            [\textbf{Nonfinite Verb Form Error:} Incorrect use of nonfinite verb forms\, such as infinitives\, gerunds\, present participles\, and past participles., inter_word_type]
            [\textbf{Verb Voice Error:} Incorrect use of active or passive voice., inter_word_type]
            [\textbf{Verb Tense Error:} The tense of the verb does not match the context., inter_word_type]
            [\textbf{Verb Choice Error:} Using inappropriate verbs or omitting necessary verbs., inter_word_type]
            [\textbf{POS Confusion Error:} Using the wrong part of speech for a word with the same root or affix., inter_word_type]
            [\textbf{Conjunction Error:} Using inappropriate conjunctions or omitting necessary conjunctions., inter_word_type]
            [\textbf{Relative Word Error:} Using inappropriate relative words or omitting necessary relative words., inter_word_type]
            [\textbf{Sentence Structure Error:} Confusing sentence structure or non-standard sentence patterns that lead to unclear meaning., inter_word_type]
            [\textbf{Word Order Error:} The arrangement of words does not conform to idiomatic expressions or grammatical rules., inter_word_type]
            [\textbf{Word Redundancy Error}: The sentence contains redundant elements that can be removed without affecting its completeness or meaning\, making it more concise and fluent., inter_word_type]
        ]
        [\textbf{Discourse Level Error}, discourse
            [\textbf{Punctuation Error:} Misuse or omission of punctuation marks in writing.
            , discourse_type]
            [\textbf{Format Error:} The text uses a format that does not conform to standard norms\, such as in letters., discourse_type]
            [\textbf{Sentence Redundancy Error}: A sentence repeats information from the previous text\, and removing it makes the text more concise., discourse_type]
        ]
    ]
\end{forest}}
\caption{Taxonomy of Error Analysis for English Writing.}
\label{Fig:taxonomy}
\end{figure*}

\subsection{Error Type Taxonomy}
\label{app:taxonomy}
We illustrate our constructed error type taxonomy in Figure~\ref{Fig:taxonomy}. We stipulate the priority of error types according to their top-to-bottom positions in Figure~\ref{Fig:taxonomy}. For instance, Case Error is assigned the lowest priority, while Sentence Redundancy Error holds the highest. In particular, Punctuation Error is prioritized between Contraction Error and Determiner Error due to its ubiquity. Therefore, when encountering an edit involving a Punctuation Error and a Determiner Error, models should classify it as a Determiner Error.

Definitions and examples of all error types in our proposed taxonomy are listed as follows:

\begin{itemize}
\item [(01)] Case Error: Incorrect use of uppercase or lowercase letters. 
Example: Writing ``paris'' instead of ``Paris.''

\item [(02)] Space Error: Missing necessary spaces between words or having extra spaces.
Example: Writing ``tothelibrary'' instead of ``to the library.''

\item [(03)] Spelling Error: The spelling of a word does not conform to standard norms. Note that both British and American spellings are considered correct and should not be classified as spelling errors.
Example: Writing ``recieve'' instead of ``receive''; ``definately'' instead of ``definitely.''

\item [(04)] Contraction Error: Incorrect use of word contractions.
Example: Writing ``isnt'' instead of ``isn't.''

\item [(05)] Punctuation Error: Misuse or omission of punctuation marks in writing. For example, two or more independent clauses are improperly joined without correct punctuation or conjunctions, or sentence components that should be joined are separated into independent sentences.
Example: Writing ``He sings children's songs he is an excellent musician'' instead of ``He sings children's songs. He is an excellent musician.''

\item [(06)] Determiner Error: Using inappropriate determiners or omitting necessary determiners, such as articles (a, an, the).
Example: Writing ``She has cat'' instead of ``She has a cat.''

\item [(07)] Number Error: Using inappropriate cardinal or ordinal numbers.
Example: Writing ``two place'' instead of ``second place.''

\item [(08)] Preposition Error: Using inappropriate prepositions or omitting necessary prepositions.
Example: Writing ``good in math'' instead of ``good at math.''

\item [(09)] Auxiliary Error: Using inappropriate auxiliary verbs or omitting necessary auxiliary verbs (including basic and modal auxiliaries).
Example: Writing ``should sing well'' instead of ``can sing well.''

\item [(10)] Adjective Error: Using inappropriate adjectives or omitting necessary adjectives, including improper use of comparative or superlative forms.
Example: Writing ``more taller'' instead of ``taller.''

\item [(11)] Adverb Error: Using inappropriate adverbs or omitting necessary adverbs.
Example: Writing ``runs quick'' instead of ``runs quickly.''

\item [(12)] Noun Number Error: Incorrect use of singular or plural forms of nouns, or confusion between countable and uncountable nouns.
Example: Writing ``many book'' instead of ``many books.''

\item [(13)] Noun Possessive Error: Incorrect use or omission of possessive forms of nouns.
Example: Writing ``Johns book'' instead of ``John's book.''

\item [(14)] Noun Error: Using inappropriate nouns or omitting necessary nouns.
Example: Writing ``The book is on the table'' instead of ``The book is on the shelf.''

\item [(15)] Pronoun Error: Using inappropriate pronouns or omitting necessary pronouns, including inconsistency in gender, number, or case.
Example: Writing ``Everyone should bring their own lunch'' instead of ``Everyone should bring his or her own lunch.''

\item [(16)] Subject-Verb Agreement Error: Mismatch between the subject and verb in number or person.
Example: Writing ``The list of items are'' instead of ``The list of items is.''

\item [(17)] Nonfinite Verb Form Error: Incorrect use of nonfinite verb forms, such as infinitives, gerunds, present participles, and past participles.
Example: Writing ``suggested to go'' instead of ``suggested going.''

\item [(18)] Verb Voice Error: Incorrect use of active or passive voice.
Example: Writing ``was ate'' instead of ``was eaten.''

\item [(19)] Verb Tense Error: The tense of the verb does not match the context.
Example: Writing ``Yesterday, I go'' instead of ``Yesterday, I went.''

\item [(20)] Verb Choice Error: Using inappropriate verbs or omitting necessary verbs.
Example: Writing ``tried to visit'' instead of ``decided to visit.''

\item [(21)] PoS Confusion Error: Using the wrong part of speech for a word with the same root or affix.
Example: Writing ``beauty singer'' instead of ``beautiful singer.''

\item [(22)] Conjunction Error: Using inappropriate conjunctions or omitting necessary conjunctions.
Example: Writing ``I wanted to go, and I was tired'' instead of ``I wanted to go, but I was tired.''

\item [(23)] Relative Word Error: Using inappropriate relative words or omitting necessary relative words.
Example: Writing ``where I was born in'' instead of ``in which I was born.''

\item [(24)] Sentence Structure Error: Confusing sentence structure or non-standard sentence patterns that lead to unclear meaning.
Example: Writing ``The book on the table which I read yesterday'' instead of ``The book which I read yesterday is on the table.''

\item [(25)] Word Order Error: The arrangement of words does not conform to idiomatic expressions or grammatical rules.
Example: Writing ``older three years'' instead of ``three years older.''

\item [(26)] Word Redundancy Error: The sentence contains redundant elements that can be removed without affecting its completeness or meaning, making it more concise and fluent.
Example: Writing ``returned back'' instead of ``returned.''

\item [(27)] Format Error: The text uses a format that does not conform to standard norms, such as in letters.
Example: Writing ``Dear Sir, I am writing to you...'' instead of ``Dear Sir,$\backslash$n I am writing to you....''

\item [(28)] Sentence Redundancy Error: A sentence repeats information from the previous text, and removing it makes the text more concise.
Example: Writing ``I went to the store. I went to the store to buy milk'' instead of ``I went to the store to buy milk.''

\item [(29)] Other Error: Errors that do not fall into the above categories.
Example: Non-sense sentences like ``I look like beauty as famous do.''
\end{itemize}

\subsection{Details of Error Severity}
\label{app:severity}

Error severity in the proposed dataset is rated from 1 (trivial) to 5 (extremely serious).

\begin{itemize}
\item 1 point (trivial): Minor issues like spelling or punctuation that don't affect understanding.

Example: "I have a \Red{friand} who likes football." (\textit{friand} $\to$ \textit{friend})

\item 2 points (minor): Errors like verb tense or simple subject-verb disagreement that don't alter the main meaning.

Example: "He \Red{go} to school every day." (\textit{go} $\to$ \textit{goes})

\item 3 points (moderate): More complex errors not easy to understand, such as clause misuse.

Example: "This is the book that I told you about \Red{it}." (remove \textit{it})

\item 4 points (serious): Multiple issues or confusing structure that hinder understanding.

Example: "Yesterday I \Red{go store} and bought some apples." (\textit{go store} $\to$ \textit{went to the store})

\item 5 points (extremely serious): Errors that make the sentence incomprehensible, often due to serious word misuse or structural issues.

Example: "My brother \Red{are play hap}." (\textit{are play hap} $\to$ \textit{is playing happily})
\end{itemize}

\begin{figure*}[tbp!]
\centering
\begin{subfigure}[b]{0.48\textwidth} 
    \centering
    \includegraphics[width=\textwidth]{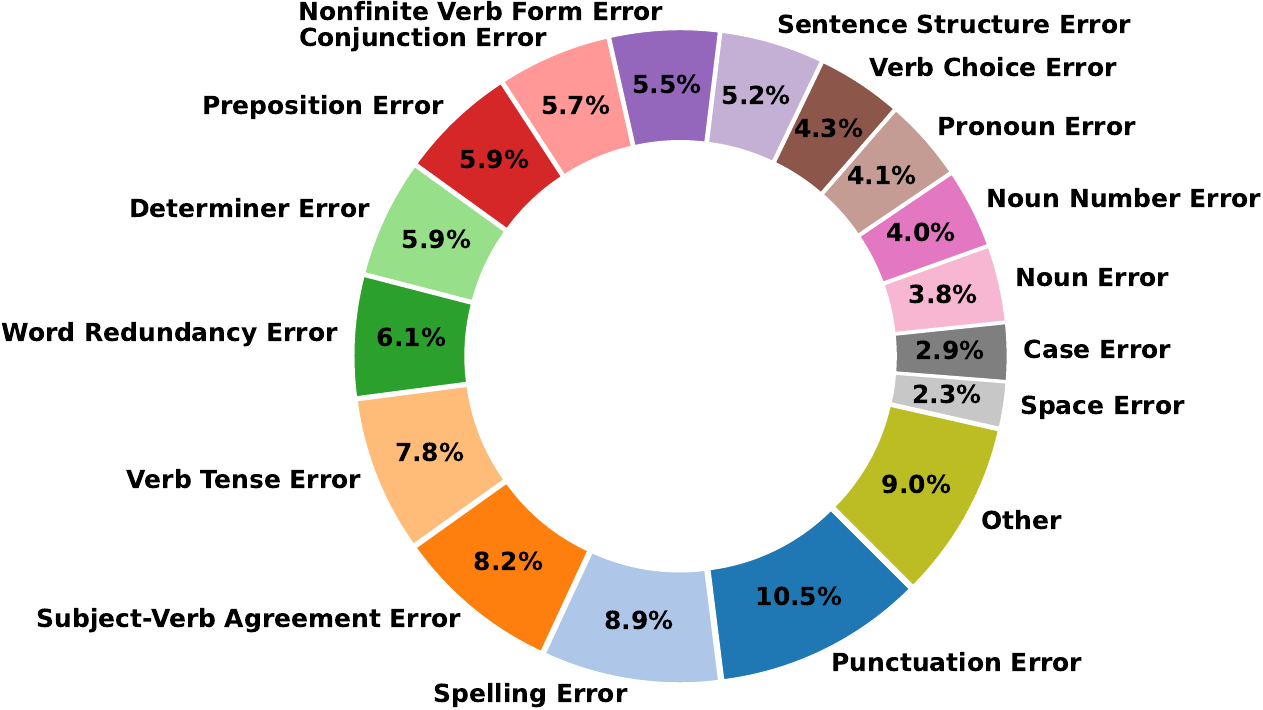}
    \caption{Error type distribution of elementary essays.}
    \label{Fig:elementary_error_type}
\end{subfigure}
\hfill 
\begin{subfigure}[b]{0.50\textwidth} 
    \centering
    \includegraphics[width=\textwidth]{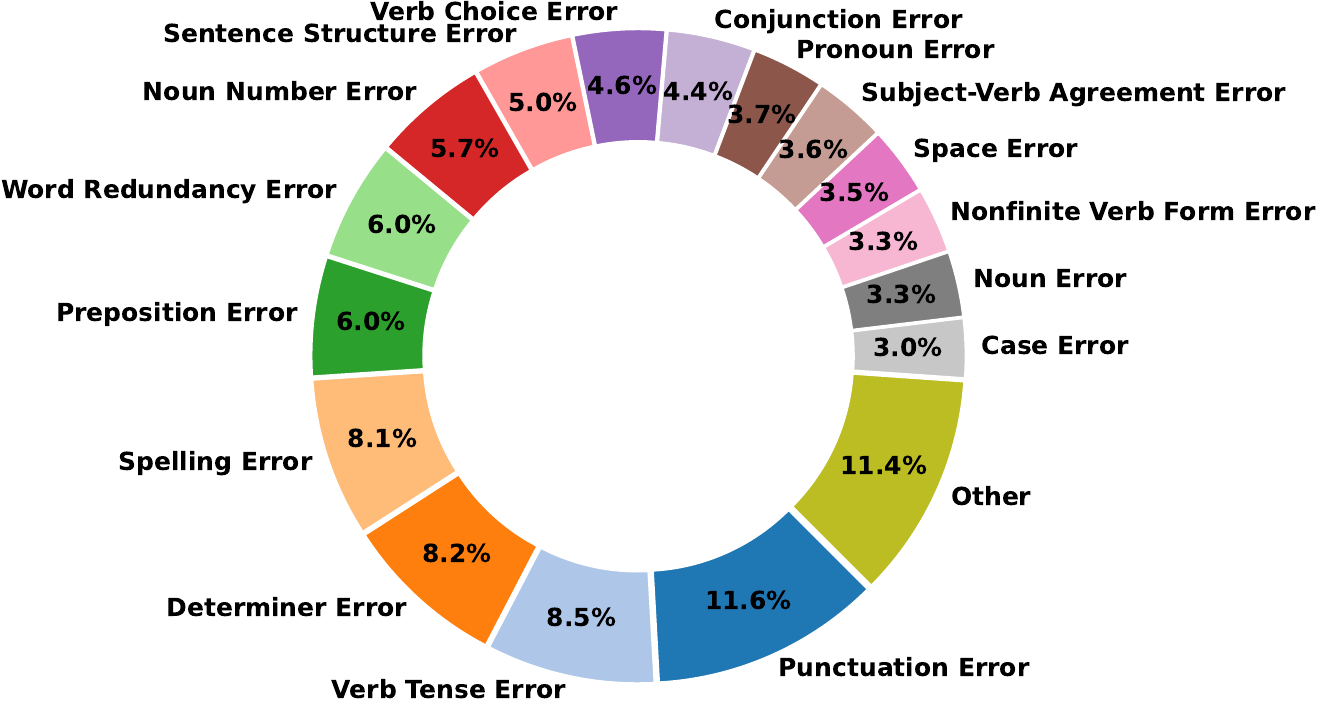} 
    \caption{Error type distribution of secondary essays.}
    \label{Fig:secondary_error_type}
\end{subfigure}
\caption{Error type distribution of \Benchmark{}. We illustrate the most frequent 16 error types out of totally 29 error types due to the space limit and present the remaining error types as ``Other''.}
\label{Fig:error_type_distribution}
\end{figure*}

\subsection{Error Type Distribution}
\label{app:error_type_distribution}
The distribution of error types across elementary and secondary school student essays in Figure~\ref{Fig:error_type_distribution} reveals that the error distributions are quite similar. Key error categories such as Punctuation Error, Spelling Error, Verb Tense Error, Word Redundancy Error, Determiner Error, and Preposition Error each account for over 5\% of the total errors observed in both elementary and secondary school essays. These represent persistent challenges for K-12 English language learners. However, notable divergences also exist. For instance, Subject-Verb Agreement errors are proportionally more prevalent in the writing of elementary school students, which is consistent with typical language acquisition trajectories. Essays from secondary school students tend to exhibit a higher proportion of errors classified under the ``Other'' category. This suggests a more pronounced long-tail effect in their error patterns, possibly due to their engagement with more complex linguistic structures and vocabulary, leading to a wider variety of less common errors.

\section{Evaluation Details and Prompts for \Benchmark{}}
\label{app:evaluation_details}

\paragraph{Evaluation Details.}
For closed-source or large models, we interact with the models through their respective APIs, ensuring consistency in input formatting and evaluation protocols. For the open-source models with parameter sizes less than 8B, we deploy them locally on NVIDIA A800 GPUs, leveraging their fine-tuned versions for conversational tasks. Each model is evaluated using identical prompts and settings to ensure fair comparisons. We set the \texttt{temperature} to 0.6 and \texttt{top\_p} to 0.95.

\paragraph{Prompts.}
Our designed prompts are shown in Figure~\ref{fig:prompt_zero_shot_naive} and Figure~\ref{fig:prompt_one_shot_detailed}.

\begin{figure*}
\begin{tcolorbox}[
    colback=gray!10,       
    colframe=darkgray,     
    boxrule=1pt,           
    arc=4pt,               
    boxsep=5pt,            
    left=6pt, right=6pt,   
    top=6pt, bottom=6pt,    
    title=Prompt for the \texttt{Zero-shot-naive} setting,
]
You are an experienced English K-12 teacher, specializing in providing accurate and relevant educational feedback for writing errors in essays.
To ensure accuracy and relevance, adhere to these principles:
\\
1. Analyze each given edit one by one. Maintain the exact number of edits and make sure the edit index is correct.
\\
2. Don't alter the error and correction content in any case.
\\
3. Specify a single error type, a severity, and a description for each edit. If an edit involves multiple errors, you must predict only one error type with the highest priority order (see below). However, when describing, you must provide a detailed explanation for each error type and use numbering such as \ding{172}, \ding{173}, and semicolons to separate the descriptions of different error types.
\\
4. Error severity is rated from 1 (trivial) to 5 (extremely serious).
\\
5. The error description must target the predicted error type, highlight the violated semantic rules and relevant knowledge, and explain the given correction and its rationale.
\\
6. Use specific symbols to emphasize evidence words and correction methods: (1) Evidence words from the error sentence are enclosed in 〈〉. (2) Correction methods from the correct sentence are enclosed in [].
\\
7. Predict a single error type for an edit based on the following error taxonomy. Directly generate the name of the error type without serial number, e.g., ``Preposition Error.'' Don't generate any other error types not included in the taxonomy.
\\ \\
Error Taxonomy:
\\
(01) Case Error
\\
...
\\ \\
8. Output strictly in the following JSON format.
\\
\{JSON Format Instruction\}
\\ \\
Now you should deal with the following input and output a single JSON output.
\\
\{Input\}
\end{tcolorbox}
\caption{Prompt for the \texttt{Zero-shot-naive} setting}
\label{fig:prompt_zero_shot_naive}
\end{figure*}

\begin{figure*}
\begin{tcolorbox}[
    colback=gray!10,       
    colframe=darkgray,     
    boxrule=1pt,           
    arc=4pt,               
    boxsep=5pt,            
    left=6pt, right=6pt,   
    top=6pt, bottom=6pt,    
    title=Prompt for the \texttt{One-shot-detailed} setting,
]
You are an experienced English K-12 teacher, specializing in providing accurate and relevant educational feedback for writing errors in essays.
To ensure accuracy and relevance, adhere to these principles:
\\
1. Analyze each given edit one by one. Maintain the exact number of edits and make sure the edit index is correct.
\\
2. Don't alter the error and correction content in any case.
\\
3. Specify a single error type, a severity, and a description for each edit. If an edit involves multiple errors, you must predict only one error type with the highest priority order (see below). However, when describing, you must provide a detailed explanation for each error type and use numbering such as \ding{172}, \ding{173}, and semicolons to separate the descriptions of different error types.
\\
4. Error severity is rated from 1 (trivial) to 5 (extremely serious).
\\
- 1 point (trivial): Minor issues like spelling or punctuation that don't affect understanding.
\\
Example: "I have a friand who likes football." (friand -> friend)
\\
- 2 points (minor): Errors like verb tense or simple subject-verb disagreement that don't alter the main meaning.
\\
Example: "He go to school every day." (go -> goes)
\\
- 3 points (moderate): More complex errors not easy to understand, such as clause misuse.
\\
Example: "This is the book that I told you about it." (remove it)
\\
- 4 points (serious): Multiple issues or confusing structure that hinder understanding.
\\
Example: "Yesterday I go store and bought some apples." (go store -> went to the store)
\\
- 5 points (extremely serious): Errors that make the sentence incomprehensible, often due to serious word misuse or structural issues.
\\
Example: "My brother are play hap." (are play hap -> is playing happily)
\\
5. The error description must target the predicted error type, highlight the violated semantic rules and relevant knowledge, and explain the given correction and its rationale.
\\
6. Use specific symbols to emphasize evidence words and correction methods: (1) Evidence words from the error sentence are enclosed in 〈〉. (2) Correction methods from the correct sentence are enclosed in [].
\\
7. Predict a single error type for an edit based on the following error taxonomy. Directly generate the name of the error type without serial number, e.g., ``Preposition Error.'' Don't generate any other error types not included in the taxonomy.
\\ \\
Error Taxonomy:
\\
(01) Case Error: Incorrect use of uppercase or lowercase letters. Example: Writing ``paris'' instead of ``Paris''.
\\
...
\\ \\
8. Here provides an input and output example strictly in JSON format.
\\
\{Example\}
\\ \\
Now you should deal with the following input and output a single JSON output.
\\
\{Input\}
\end{tcolorbox}
\caption{Prompt for the \texttt{One-shot-detailed} setting}
\label{fig:prompt_one_shot_detailed}
\end{figure*}

\section{Extra Results}
\label{app:extra_results}

\begin{table*}[tbp!]
\centering
\renewcommand{\arraystretch}{1.2}
\renewcommand{\tabcolsep}{3pt}
\resizebox{0.95\linewidth}{!}{
\begin{tabular}{lccccccccccccc}
\toprule

\multirow{2}{*}{\textbf{Setting}}
& \multirow{2}{*}{\textbf{Pre-exp.}}


& \multicolumn{4}{c}{\textbf{Classification}}
& \multicolumn{2}{c}{\textbf{Severity}}
& \multicolumn{6}{c}{\textbf{Explanation}}
\\
\cmidrule(lr){3-6} \cmidrule(lr){7-8} \cmidrule(ll){9-14}

&
& \multicolumn{2}{c}{\textbf{Acc}$\uparrow$}
& \multicolumn{2}{c}{\textbf{F$_1$}$\uparrow$}
& \multicolumn{2}{c}{\textbf{MAE}$\downarrow$}
& \multicolumn{2}{c}{\textbf{BLEU}$\uparrow$} 
& \multicolumn{2}{c}{\textbf{METEOR}$\uparrow$}
& \multicolumn{2}{c}{\textbf{ROUGE}$\uparrow$} \\

\midrule

\multirow{2}{*}{\textbf{Zero-shot-naive}}
& \ding{55} & 61.74 & \cellcolor{softgreen} 66.79 & 46.55 & \cellcolor{softgreen} 52.67 & 0.87 & \cellcolor{softgreen} 0.77 & \bf 1.35 & \cellcolor{softgreen} \bf 1.19 & 17.87 & \cellcolor{softgreen} \bf 17.58 & \bf 24.29 & \cellcolor{softgreen} \bf 23.02 \\

& \ding{51} & \bf 62.04 & \cellcolor{softgreen} \bf 67.56 & \bf 48.04 & \cellcolor{softgreen} \bf 54.51 & \bf 0.85 & \cellcolor{softgreen} \bf 0.76 & 1.31 & \cellcolor{softgreen} 1.07 & 17.97 & \cellcolor{softgreen} 17.32 & 24.37 & \cellcolor{softgreen} 22.99 \\

\midrule





\multirow{2}{*}{\textbf{One-shot-detailed}}

& \ding{55} & \bf 66.57 & \cellcolor{softgreen} 72.02 & \bf 52.97 & \cellcolor{softgreen} 57.66 & 0.78 & \cellcolor{softgreen} \bf 0.66 & 2.46 & \cellcolor{softgreen} \bf 2.16 & 19.96 & \cellcolor{softgreen} 19.70 & 28.41 & \cellcolor{softgreen} 26.13 \\

& \ding{51} & 65.90 & \cellcolor{softgreen} \bf 72.62 & 51.46 & \cellcolor{softgreen} \bf 59.33 & \bf 0.70 & \cellcolor{softgreen} \bf 0.66 & \bf 2.61 & \cellcolor{softgreen} 2.05 & \bf 20.42 & \cellcolor{softgreen} \bf 19.76 & \bf 29.14 & \cellcolor{softgreen} \bf 26.50 \\

\bottomrule
\end{tabular}}
\caption{Ablation results of prediction order. We color the results of the \colorbox{softgreen}{secondary} domain.}
\label{Tab:ablation_prediction_order}
\end{table*}

\begin{figure*}[ht]
\centering
\includegraphics[width=1.0\textwidth]{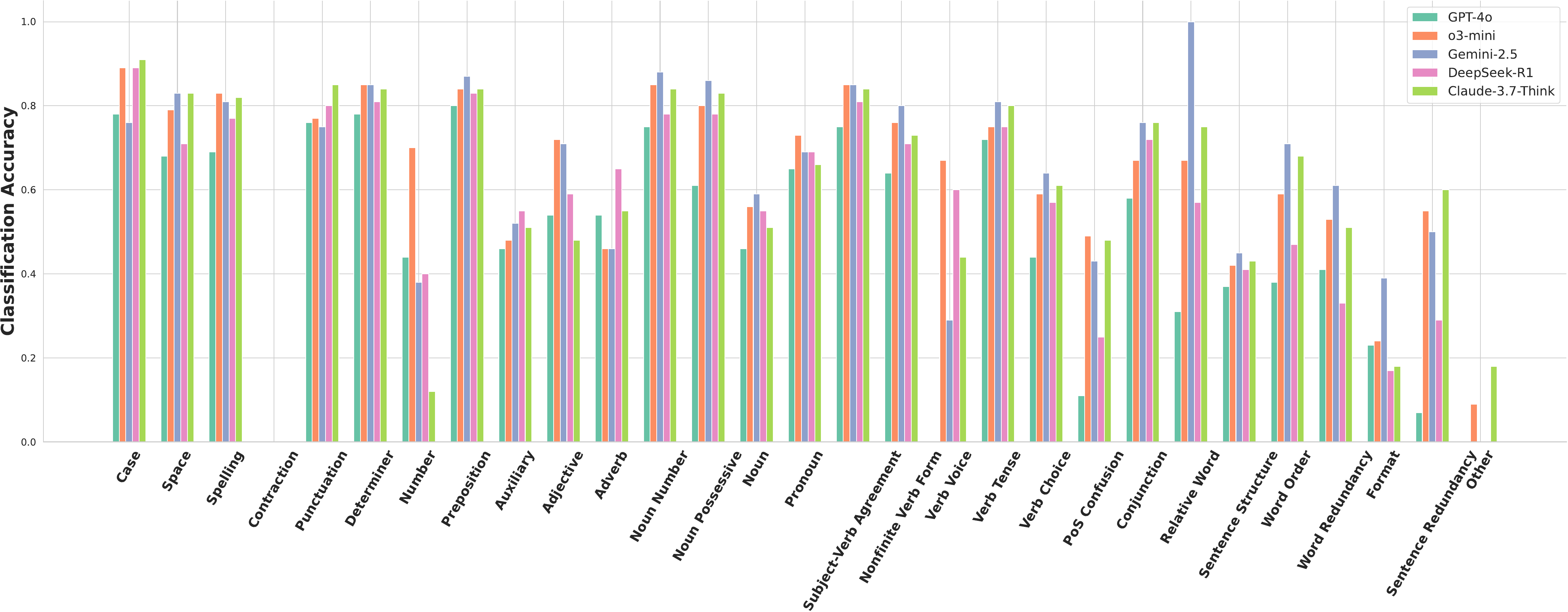}
\caption{Classification accuracy of GPT-4o, o3-mini, Gemini-2.5-pro, DeepSeek-R1, and Claude-3.7-Think. We list the accuracy of all 29 error types.}
\label{Fig:classification}
\end{figure*}

\subsection{Performance on Each Error Type}
\label{subsec:performance_error_type}
A more granular examination of model performance across individual error categories is shown in Figure~\ref{Fig:classification}. LLMs generally achieve high classification accuracy on frequent and structurally simple error types such as Case Error, Space Error, and Spelling Error. However, their performance significantly degrades on less frequent or long-tail categories and those requiring deeper linguistic understanding or contextual reasoning. These challenging types include Contraction Error, Number Error, Auxiliary Verb Error, Part-of-Speech (PoS) Confusion Error, Sentence Structure Error, and Format Error. This disparity underscores a key deficiency in current LLMs: an incomplete or insufficiently nuanced grasp of the full spectrum of error types defined within our comprehensive taxonomy, particularly those that are either rare in typical training data or inherently more complex and semantically subtle, calling for future improvement.

\subsection{Effect of Prediction Order}
\label{subsec:ablation_prediction_order}
To investigate the influence of the generation sequence on model performance, we conducted an ablation study by altering the prediction order of the three main components. Our default approach, termed \textit{Post-explaining}, predicts elements in the sequence: Error Severity $\to$ Error Type $\to$ Error Explanation. We compared this against an alternative, \textit{Pre-explaining}, which follows the order: Error Explanation $\to$ Error Severity $\to$ Error Type. All experiments for this ablation were performed using the GPT-4o model, and the results are detailed in Table~\ref{Tab:ablation_prediction_order}.
In the \texttt{Zero-shot-naive} setting, the \textit{Pre-explaining} order (generating the explanation before the error type) leads to improved performance in Error Classification and Error Severity Rating. However, this comes at the cost of reduced quality in the Error Explanation itself. This observation aligns with the intuition that outputs generated earlier in a sequence can serve as valuable contextual information for subsequent steps. Interestingly, this trend reverses in the \texttt{One-shot-detailed} setting, where the \textit{Pre-explaining} model demonstrates nearly superior performance across all three aspects. We attribute this significant shift to the influence of the in-context demonstration. The demonstration likely provides a strong template or implicit guidance on how to structure the thought process when generating an explanation first effectively, and then coherently deriving the error type and severity from that explanation.

\end{document}